%% file: root.tex
\documentclass[lettersize,journal]{IEEEtran}
\usepackage{amsmath,amsfonts}
\usepackage{algorithm}
\usepackage{amssymb}
\usepackage{array}
\usepackage{textcomp}
\usepackage{stfloats}
\usepackage{url}
\usepackage{verbatim}
\usepackage{graphicx}
\usepackage{cite}
\usepackage[hidelinks]{hyperref}
\usepackage{multirow}
\usepackage{booktabs}
\usepackage{subcaption}
\usepackage{makecell}
\usepackage{algpseudocode}
\usepackage{newtxtext}
\usepackage{newtxmath}
\usepackage[cal=cm]{mathalfa}
\usepackage[table]{xcolor}
\usepackage[top=20.1mm, bottom=15.2mm, left=16.9mm, right=16.9mm]{geometry}
\begin{document}

\title{Learning from Demonstration with Failure Awareness for Safe Robot Navigation}

\author{Xianghui Wang$^{1,2}$, Siwei Cheng$^{2,3}$, Shanze Wang$^{2,4}$, Xinming Zhang$^{2,3}$, Dan Zhang$^{1}$, Wei Zhang$^{2,\dagger}$%

\thanks{This work has been submitted to the IEEE for possible publication. Copyright may be transferred without notice, after which this version may no longer be accessible.}

\thanks{*This work was supported by the National Natural Science Foundation of China (Grant No. 62503251).}

\thanks{$^{1}$The authors are with Department of Mechanical Engineering, The Hong Kong Polytechnic University, Hung Hom, Kowloon, Hong Kong. 
        {\tt\small carlos.wang@connect.polyu.hk,\ dan.zhang@polyu.edu.hk}}%
\thanks{$^{2}$The authors are with College of Information Science and Technology, Eastern Institute of Technology, Ningbo, China. 
        {\tt\small \{carlos.wang,shanze.wang\}@connect.polyu.hk, zhw@eitech.edu.cn, \{chengsiwei, xm\_zhang\}@mail.ustc.edu.cn}}%
\thanks{$^{3}$The authors are with School of Computer Science and Technology, University of Science and Technology of China, Hefei, China. 
        {\tt\small \{chengsiwei, xm\_zhang\}@mail.ustc.edu.cn}}%
\thanks{$^{4}$The author is with Department of Aeronautical and Aviation Engineering, The Hong Kong Polytechnic University, Hung Hom, Kowloon, Hong Kong. 
        {\tt\small shanze.wang@connect.polyu.hk}}%
\thanks{$\dagger$ is the corresponding author.}
}



\maketitle

\begin{abstract}
Learning from demonstration is widely used for robot navigation, yet it suffers from a fundamental limitation: demonstrations consist predominantly of successful behaviors and provide limited coverage of unsafe states. This limitation leads to poor safety when the robot encounters scenarios beyond the demonstration distribution. Failure experiences, such as collisions, contain essential information about unsafe regions, but remain underutilized. The key difficulty lies in the fact that failure data do not provide valid guidance for action imitation, and their naive incorporation into policy learning often degrades performance. We address this challenge by proposing a failure-aware learning framework that explicitly decouples the roles of success and failure data. In this framework, failure experiences are used to shape value estimation in hazardous regions, while policy learning is restricted to successful demonstrations. This separation enables the effective use of failure data without corrupting policy behavior. We implement this design within an offline reinforcement learning (RL) setting and evaluate it in both simulation and real-world environments. The results show that our framework consistently reduces collision rates while preserving the task success rate, and demonstrate strong generalization across different environments and robot platforms.
\end{abstract}


\input{Chapters/Introduction}

\input{Chapters/Related_Work}

\input{Chapters/Preliminaries}

\input{Chapters/Approach}

\input{Chapters/Experiment}

\input{Chapters/Conclusion}


\bibliographystyle{IEEEtran}
\bibliography{mybib}

\end{document}

%% file: Chapters/Introduction.tex
\section{Introduction}
Learning from demonstration has become a widely adopted paradigm for mapless robot navigation due to its ability to learn effective policies directly from expert data~\cite{zare2024survey}. However, it suffers from a fundamental limitation: expert demonstrations consist predominantly of successful trajectories and provide limited coverage about unsafe behaviors~\cite{yan2022mapless,torabi2018behavioral}. As a result, policies trained on demonstration data often lack awareness of failure experience, leading to an increased collision risk when encountering states outside the coverage of the expert demonstrations~\cite{ross2011reduction}. This limitation is particularly critical for safety-sensitive navigation tasks, where avoiding rare but catastrophic events is essential.

Existing approaches attempt to address this limitation from different perspectives. Imitation learning methods primarily rely on expert demonstrations and are thus inherently limited by their lack of failure coverage~\cite{ross2011reduction}. Techniques such as Dataset Aggregation (DAgger) and Generative Adversarial Imitation Learning (GAIL) attempt to mitigate this issue by querying additional supervision or matching expert behavior~\cite{ross2011reduction,ho2016generative}, but they remain heavily dependent on expert data and fail to explicitly utilize failure experiences. In contrast, offline reinforcement learning (RL) offers the ability to learn from suboptimal data, making it a promising framework for incorporating failure experiences~\cite{kumar2022should}. However, naively mixing successful and failure data for offline RL training often results in unstable or degraded policies~\cite{hoang2025no}.

The key challenge lies in the fundamentally different roles of success and failure data in policy learning. Specifically, successful demonstrations provide valid action guidance for reaching the goal, whereas failure experiences primarily convey information about unsafe regions rather than desirable actions. Consequently, treating these two types of data uniformly leads to performance degradation of navigation safety.

To address this challenge, we propose a failure-aware learning framework that explicitly decouples the roles of success and failure data. The core idea is to incorporate failure experiences in a structured manner so as to improve navigation safety without compromising policy learning. This design enables effective utilization of failure experiences without introducing undesirable behaviors into the learned policy.

We instantiate this framework within an offline RL setting and develop a structured learning approach for safe robot navigation. \textbf{Rather than introducing an incremental modification to offline RL, our method provides a failure-aware perspective on learning from demonstration.} At the data level, success and failure transitions are separated and rebalanced during training. At the algorithm level, an asymmetric update mechanism is employed, where failure data are leveraged to shape value estimation, while policy updates are guided solely by success transitions.

Extensive experiments in both simulation and real-world environments demonstrate that the proposed framework significantly reduces collision rates while maintaining high success rates. Moreover, the learned policy generalizes across environments and robot platforms, highlighting the effectiveness of leveraging failure experiences in a principled manner. In summary, the main contributions of this work are as follows:

\begin{enumerate}
    \item We identify a fundamental limitation of learning from demonstration for robot navigation: the lack of failure coverage in expert data leads to poor safety generalization.
    \item We propose a failure-aware learning framework to incorporate failure experiences into policy learning by explicitly decoupling their roles from successful demonstrations.
    \item We introduce a structured learning approach that leverages failure data for value estimation while preserving policy learning on successful demonstration.
    \item We validate the proposed framework through extensive simulation and real-world experiments, demonstrating significant improvements in navigation safety without sacrificing task performance.
\end{enumerate}

%% file: Chapters/Related_Work.tex
\section{Related Work}

\subsection{Learning Navigation from Expert Demonstration}
In recent years, IL-based methods have shown strong performance in mapless navigation by directly learning end-to-end policies from expert demonstrations~\cite{Kim2015DeepNN}. Early work explored IL-based end-to-end policies using image or laser observations for collision avoidance~\cite{sergeant2015multimodal,muller2005off}. However, these studies did not consider navigation tasks with explicit goal-reaching objectives. Subsequent IL-based navigation studies incorporated relative goal information into policy learning, thereby enabling target-reaching navigation~\cite{pfeiffer2017perception,pfeiffer2018reinforced}. However, these methods remain sensitive to distributional shift, as expert demonstrations rarely cover near-obstacle scenarios, often resulting in failed recovery or even collisions. To address this, DAgger was proposed to iteratively query expert actions on states visited by the current policy, thereby helping mitigate distributional shift. In addition, GAIL uses a discriminator to align the learner's behavior with that of the expert, which helps alleviate distributional shift~\cite{ho2016generative}. However, these methods remain constrained by expert demonstrations, leaving their performance typically upper-bounded by the demonstrator~\cite{brown2020better}.

\subsection{Offline Reinforcement Learning}
Offline RL refers to data-driven approaches that optimize policies from pre-collected datasets, and its ability to learn from suboptimal datasets offers the potential for more generalized and superior policies~\cite{kumar2019stabilizing,Fujimoto2018OffPolicyDR}. Prior works can broadly be divided into two categories. The first is policy-constraint methods, which regularize the learned policy to stay close to the behavior policy explicitly or implicitly~\cite{wu2019behavior,liu2020provably}. The second is conservative methods, which optimize the policy against a conservative estimate of return~\cite{kumar2020conservative}. Our goal is not to devise a new offline RL algorithm. Instead, we use offline RL as a principled framework to study how collision samples can be utilized in mapless navigation to overcome the limitation of expert-only imitation learning and improve safety. Prior robotic applications mainly use offline RL itself for navigation, with representative results largely reported in outdoor settings~\cite{shah2022offline,weerakoon2024vapor}. In contrast, we focus on using it to exploit collision samples in a structured manner for indoor mapless navigation, so as to reduce collisions without affecting navigation performance.

%% file: Chapters/Preliminaries.tex
\section{Preliminaries}
\label{sec:problem_formulation}

\begin{figure}[t]
    \centering
    \includegraphics[width=0.5\linewidth]{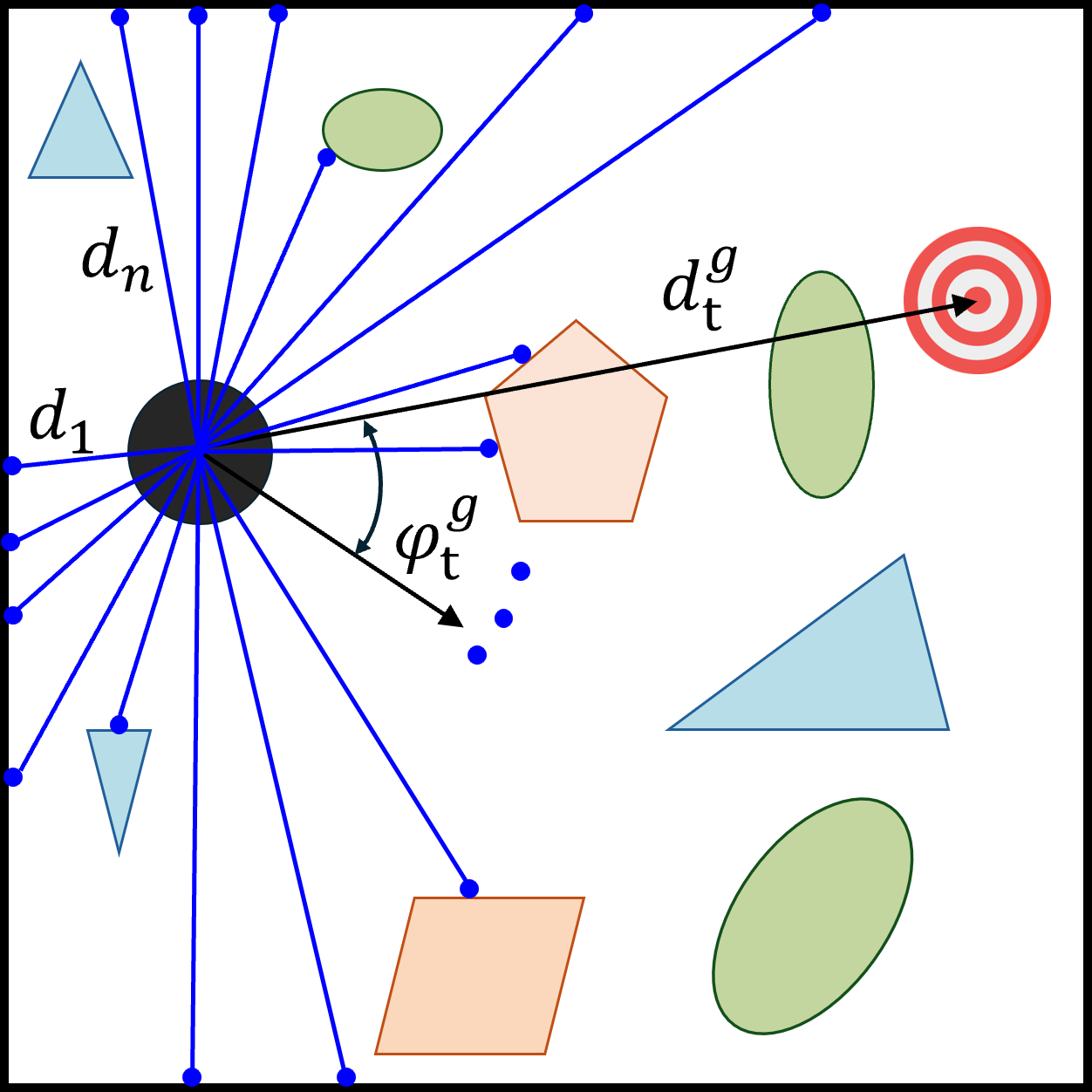}
    \caption{Illustration of the mapless robot navigation problem.}
    \label{fig:problem_illustration}
\end{figure}

\subsection{Problem Formulation}
Mapless robot navigation is a sequential decision-making problem. As shown in Fig.~\ref{fig:problem_illustration}, a mobile robot equipped with a 2--D LiDAR must reach a target position in an unknown environment while avoiding collisions with surrounding obstacles. Therefore, this task can be formulated as a Markov decision process (MDP), $\mathcal{M} = (\mathcal{S}, \mathcal{A}, T, r, \gamma)$, where $\mathcal{S}$ and $\mathcal{A}$ denote the state and action spaces, $T$ denotes the transition dynamics, $r$ denotes the reward function, and $\gamma \in [0,1)$ is the discount factor. At time step $t$, the state is defined as $s_t = \{s_t^o, s_t^g, v_t, \omega_t\}$, where $s_t^o$ denotes the LiDAR measurements, and $v_t$ and $\omega_t$ denote the current linear and angular velocities, respectively. We assume that the relative goal position in the robot coordinate frame, $s_t^g = \{d_t^g, \varphi_t^g\}$, can be obtained through localization techniques~\cite{chen2016achieving}. The action is $a_t = \{v_t^{\mathrm{cmd}}, \omega_t^{\mathrm{cmd}}\}$, which denotes the commanded linear and angular velocities. The reward $r_t$ is defined as follows:
\begin{equation}
    r_t =
    \begin{cases}
        r_s, & \text{if success},\\
        r_c, & \text{if collision},\\
        c_1 \left(d_t^g - d_{t+1}^g\right), & \text{otherwise}.
    \end{cases}
    \label{eq:reward}
\end{equation}
where $r_s$ is awarded upon goal reaching, $r_c$ penalizes collision, and otherwise a small dense reward term is provided to encourage the robot to move toward the goal, with $c_1$ being the scaling constant.

\subsection{Offline Dataset with Success and Failure Transitions}

In the offline setting, the policy is learned from a pre-collected dataset without further interaction with the environment. The dataset is defined as:
\begin{equation}
\mathcal{D} = \{(s, a, r, s')\},
\end{equation}

In this work, we explicitly distinguish two types of transitions within the dataset:
\begin{equation}
\mathcal{D} = \mathcal{D}_{\text{exp}} \cup \mathcal{D}_{\text{col}},
\end{equation}
where $\mathcal{D}_{\text{exp}}$ and $\mathcal{D}_{\text{col}}$ consist of transitions from successful and collision trajectories, respectively.

This distinction is essential since these transitions play distinct roles in policy learning. Specifically, transitions in $\mathcal{D}_{\text{exp}}$ provide valid action guidance, whereas those in $\mathcal{D}_{\text{col}}$ primarily represent unsafe regions rather than desirable behaviors. Treating these transitions uniformly often leads to the degradation of navigation safety.

\subsection{Offline Reinforcement Learning Objective}

The objective of offline RL is to learn a policy $\pi(a|s)$ that maximizes the expected discounted cumulative return. The discounted return at time step $t$ is defined as follows:
\begin{equation}
R_t = \sum_{k=t}^{H-1} \gamma^{k-t} r_k.
\end{equation}

Unlike online RL, the agent cannot perform further environment interactions to correct its behavior and must rely entirely on the pre-collected dataset $\mathcal{D}$. Consequently, the learned policy is highly sensitive to the specific composition of the transitions in $\mathcal{D}$.

This dependency makes the integration of failure experiences particularly challenging. Without the capability for online exploration, treating successful and failure transitions uniformly during optimization may cause the policy to become overly conservative or lead to an unintended bias toward unsafe behaviors, thereby compromising navigation safety.

\subsection{Implicit Q-Learning}

\begin{figure}[t]
    \centering
    \includegraphics[width=0.98\linewidth]{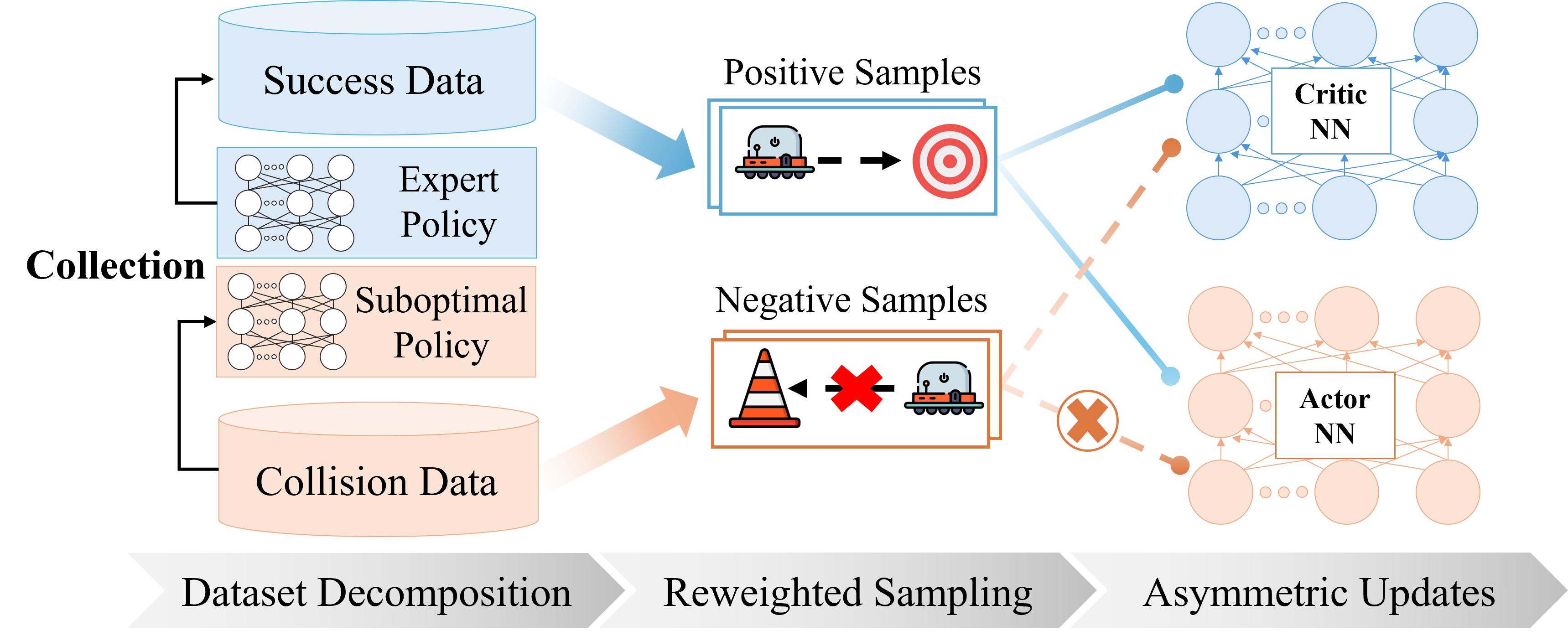}
    \caption{Overall framework of the proposed method}
    \label{fig:method_framework}
\end{figure}

To instantiate our framework, we adopt Implicit Q-Learning (IQL) as the offline RL backbone due to its strong empirical performance. IQL mitigates extrapolation errors by avoiding explicit out-of-distribution (OOD) queries~\cite{kostrikov2021iql}. Specifically, IQL optimizes the following objectives:

\begin{equation}
\mathcal{L}_V(\psi) = \mathbb{E}_{(s,a)\sim\mathcal{D}} 
\left[ L_{\tau}^2 \big( Q_{\hat{\theta}}(s,a) - V_{\psi}(s) \big) \right],
\end{equation}
\begin{equation}
\mathcal{L}_Q(\theta) = \mathbb{E}_{(s,a,s')\sim\mathcal{D}} 
\left[ \big( r(s,a) + \gamma V_{\psi}(s') - Q_{\theta}(s,a) \big)^2 \right],
\end{equation}
where $L_{\tau}^2(\cdot)$ denotes the asymmetric squared loss used in expectile regression, and $\tau$ is the expectile parameter.

In addition, the policy is subsequently extracted via advantage-weighted regression by minimizing:
\begin{equation}
\mathcal{L}_\pi(\phi) = - \mathbb{E}_{(s,a)\sim\mathcal{D}}
\left[ \log \pi_\phi(a|s) \cdot \exp(\beta A(s,a)) \right],
\end{equation}
where $A(s,a) = Q_{\hat{\theta}}(s,a) - V_{\psi}(s)$, and $\beta$ is a temperature parameter. 

In this work, our goal is not to modify the offline RL algorithm itself, but to develop a principled framework for the structured use of different types of transitions within the dataset.

%% file: Chapters/Approach.tex
\section{Approach}

\begin{algorithm}[t]
\caption{Asymmetric IQL with Reweighted Sampling}
\label{alg:asym_iql}
\begin{algorithmic}[1]
\Require Offline datasets $\mathcal{D}_{\mathrm{exp}}$ and $\mathcal{D}_{\mathrm{col}}$
\Require Collision sampling ratio $\rho$, batch size $B$
\Require Step sizes $\lambda_V,\lambda_Q,\lambda_\pi$, target update coefficient $\alpha$
\State Initialize value parameters $\psi$, critic parameters $\theta$, actor parameters $\phi$
\State Initialize target critic parameters $\hat{\theta} \gets \theta$
\For{each training step}
    \State Sample $\mathcal{B}_{\mathrm{col}} \sim \mathcal{D}_{\mathrm{col}}$ with size $\rho B$
    \State Sample $\mathcal{B}_{\mathrm{exp}}^{Q} \sim \mathcal{D}_{\mathrm{exp}}$ with size $(1-\rho)B$
    \State $\mathcal{B}_{\mathrm{mix}} \gets \textsc{Merge}(\mathcal{B}_{\mathrm{exp}}^{Q}, \mathcal{B}_{\mathrm{col}})$
    \State $\psi \leftarrow \psi - \lambda_V \nabla_\psi \mathcal{L}_V(\psi;\mathcal{B}_{\mathrm{mix}})$
    \State $\theta \leftarrow \theta - \lambda_Q \nabla_\theta \mathcal{L}_Q(\theta;\mathcal{B}_{\mathrm{mix}})$
    \State $\hat{\theta} \leftarrow (1-\alpha)\hat{\theta} + \alpha\theta$
    \State Sample $\mathcal{B}_{\mathrm{exp}}^{\pi} \sim \mathcal{D}_{\mathrm{exp}}$
    \State $\phi \leftarrow \phi - \lambda_\pi \nabla_\phi \mathcal{L}_\pi(\phi;\mathcal{B}_{\mathrm{exp}}^{\pi})$
\EndFor
\State \Return $\pi_\phi$
\end{algorithmic}
\end{algorithm}

\subsection{Failure-Aware Learning Principle}

We begin by analyzing the role of different types of data in offline navigation learning. 
Given an offline dataset $\mathcal{D} = \mathcal{D}_\text{exp} \cup \mathcal{D}_\text{col}$, transitions in $\mathcal{D}_\text{exp}$ correspond to successful behaviors and provide valid supervision for policy learning, while transitions in $\mathcal{D}_\text{col}$ arise from collision trajectories and indicate unsafe regions of the state-action space.

A key observation is that these two types of data play fundamentally different roles. Successful transitions describe desirable actions that should be imitated, whereas failure transitions do not provide valid action supervision and instead convey information about what should be avoided. Based on this, we adopt the following principle:

\textbf{Failure-aware learning principle:}Failure data should not be directly used for policy supervision, but should influence policy learning indirectly through value estimation. 

Under this principle, successful transitions guide policy learning, while failure transitions shape the value function by assigning lower values to unsafe regions. This allows the learned policy to avoid unsafe behaviors without explicitly imitating them.

\subsection{Framework Overview}

Guided by the failure-aware learning principle, as illustrated in Fig.~\ref{fig:method_framework}, we design a structured learning framework that decouples the roles of success and failure data. The framework consists of three key components: dataset decomposition, reweighted sampling, and asymmetric updates, which together enable the effective utilization of failure data while preserving policy learning.

First, we decompose the offline dataset as
\begin{equation}
\mathcal{D} = \mathcal{D}_\text{exp} \cup \mathcal{D}_\text{col},
\end{equation}
where $\mathcal{D}_\text{exp}$ contains transitions from successful trajectories and $\mathcal{D}_\text{col}$ contains transitions from collision trajectories. This decomposition explicitly distinguishes desirable behaviors from unsafe experiences and forms the basis for structured learning.

Second, to control the influence of failure data during training, we introduce a reweighted sampling strategy. 
A sampling ratio $\rho$ is used to regulate the proportion of collision transitions in each mini-batch, ensuring that failure data provide sufficient safety signals without dominating the learning process.

Finally, we adopt an asymmetric update mechanism to reflect the different roles of the two datasets. 
The critic network, including the value and Q networks, is trained using both $\mathcal{D}_\text{exp}$ and $\mathcal{D}_\text{col}$, allowing failure data to shape value estimation. In contrast, the policy is updated using only $\mathcal{D}_\text{exp}$, preventing it from directly imitating undesirable actions.

Together, these components implement the proposed failure-aware learning principle in a structured manner, enabling the learned policy to avoid unsafe behaviors while maintaining strong navigation performance.

\subsection{Learning Objective}

\begin{figure}[t]
    \centering
    \begin{subfigure}[b]{0.325\columnwidth}
        \centering
        \includegraphics[width=\textwidth]{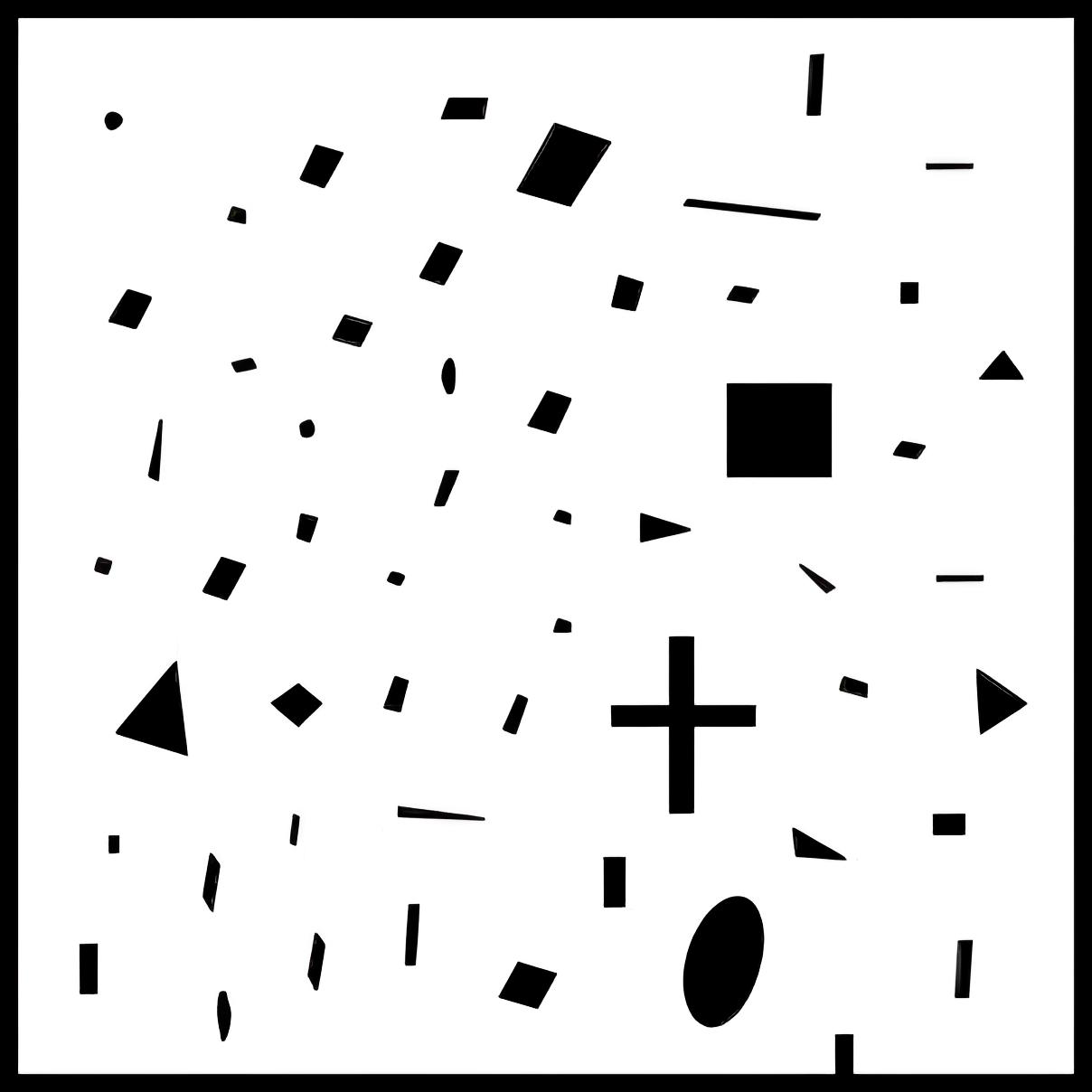}
        \caption{}
        \label{fig:SEnv1}
    \end{subfigure}
    \hfill
    \begin{subfigure}[b]{0.325\columnwidth}
        \centering
        \includegraphics[width=\textwidth]{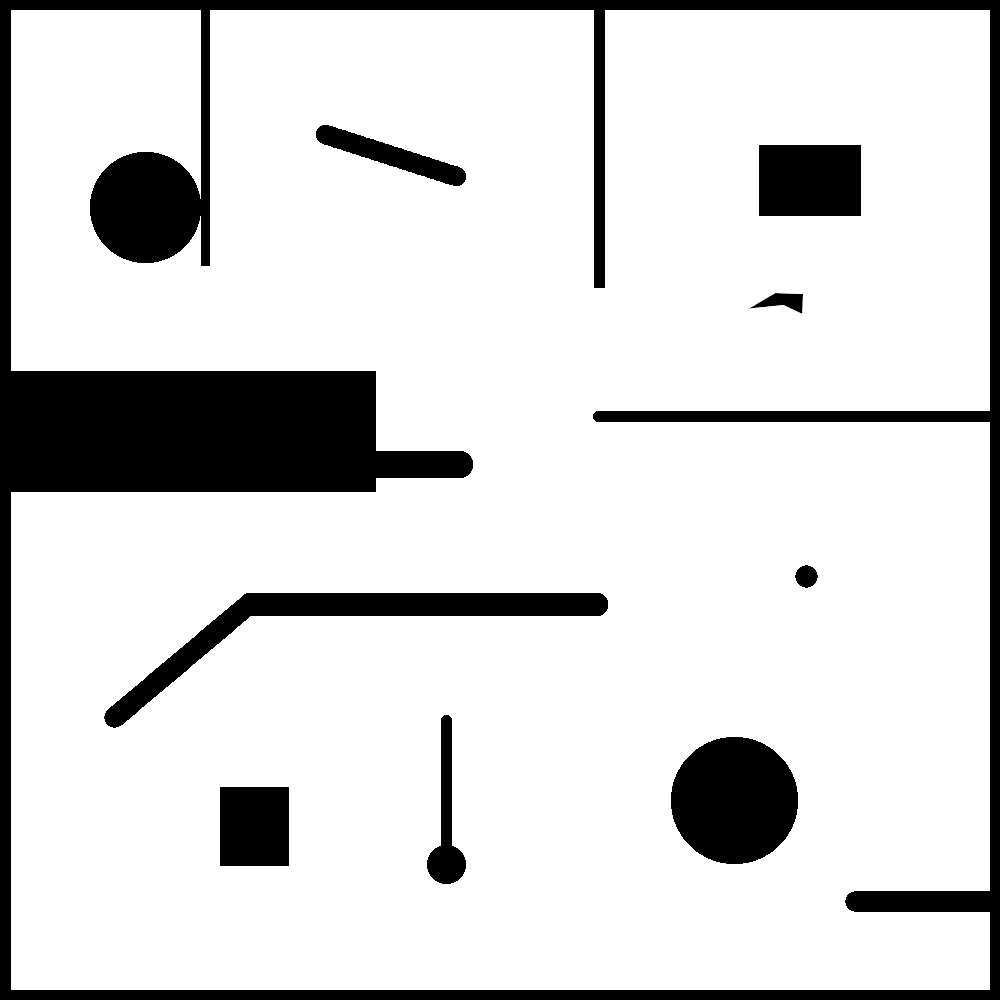}
        \caption{}
        \label{fig:SEnv2}
    \end{subfigure}
    \hfill
    \begin{subfigure}[b]{0.325\columnwidth}
        \centering
        \includegraphics[width=\textwidth]{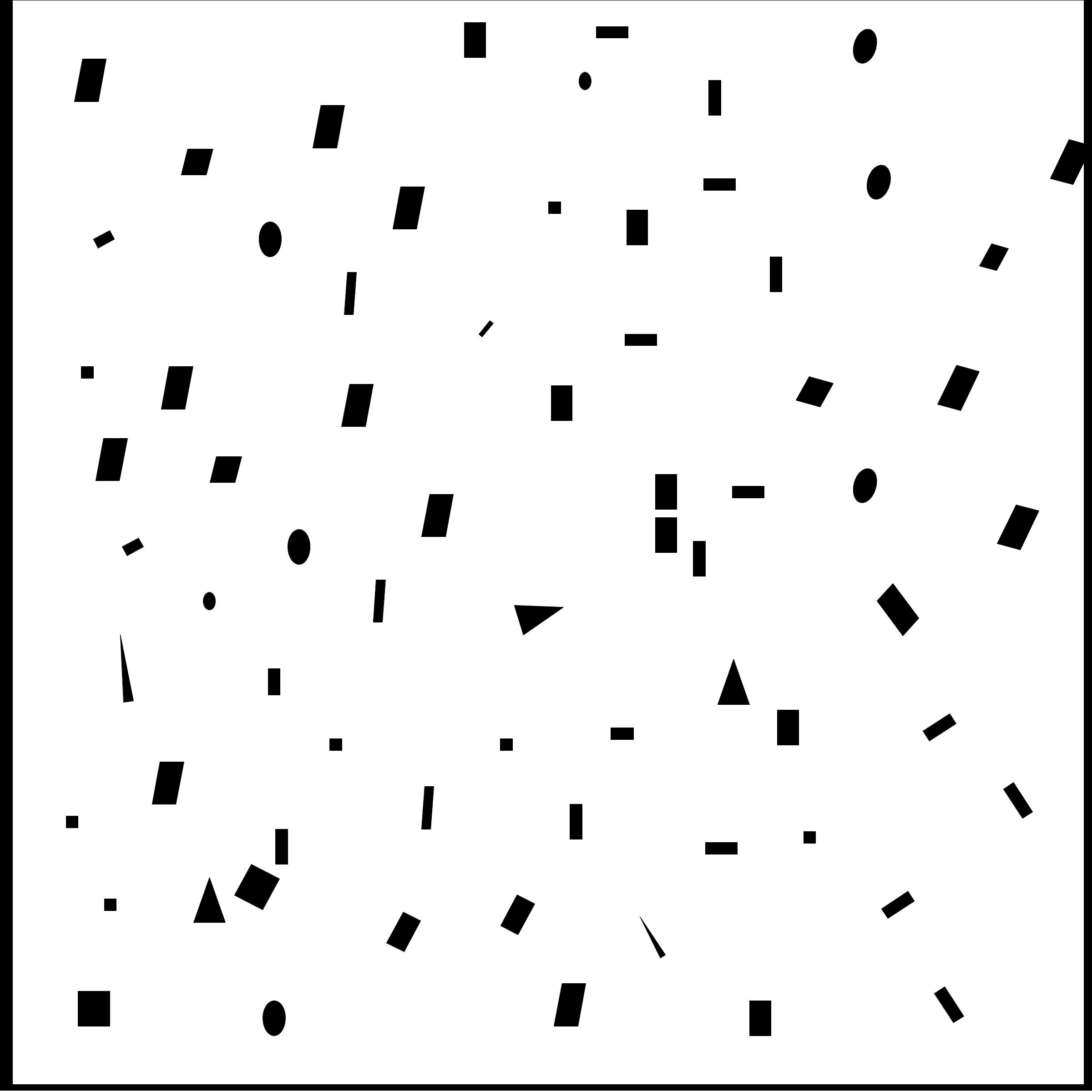}
        \caption{}
        \label{fig:SEnv3}
    \end{subfigure}
    \caption{Simulated scenarios used in the experiments. (a) SEnv1~\cite{zhang2022ipaprec}. (b) SEnv2~\cite{zhang2022ipaprec}. (c) SEnv3~\cite{zhang2021learn}.}
    \label{fig:SEnvs}
\end{figure}

We instantiate the proposed framework using IQL as the backbone algorithm. To execute the asymmetric update mechanism, the state-value function $V_\psi$ and action-value function $Q_\theta$ are optimized over the full dataset $\mathcal{D}$, whereas the policy $\pi_\phi$ is extracted exclusively from $\mathcal{D}_{\text{exp}}$.

The action-value function is trained by minimizing the temporal difference loss:
\begin{equation}
\mathcal{L}_Q(\theta) = \mathbb{E}_{(s,a,s')\sim\mathcal{D}} 
\left[
\big(r(s,a) + \gamma V_\psi(s') - Q_\theta(s,a)\big)^2
\right]
\end{equation}
where $\theta$ denotes the parameters of the action-value network.

Concurrently, the state-value function is trained using expectile regression:
\begin{equation}
\mathcal{L}_V(\psi) = \mathbb{E}_{(s,a)\sim\mathcal{D}} 
\left[
L_\tau^2\big(Q_{\hat{\theta}}(s,a) - V_\psi(s)\big)
\right]
\end{equation}
where $\psi$ and $\hat{\theta}$ represent the parameters of the state-value network and the target action-value network, respectively.

The policy is updated using advantage-weighted regression restricted to successful demonstrations:
\begin{equation}
\mathcal{L}_\pi(\phi) = - \mathbb{E}_{(s,a)\sim\mathcal{D}_\text{exp}} 
\left[
\log \pi_\phi(a|s) \cdot \exp\big(\beta A(s,a)\big)
\right]
\end{equation}
where $\phi$ denotes the parameters of the policy network, and $A(s,a) = Q_{\hat{\theta}}(s,a) - V_\psi(s)$ represents the advantage estimate.

This formulation ensures that failure data contribute to value estimation while policy learning remains guided strictly by successful demonstrations. The overall training procedure is summarized in Algorithm~\ref{alg:asym_iql}.  At each training step, a collision batch $\mathcal{B}_{\mathrm{col}}$ and an expert batch $\mathcal{B}_{\mathrm{exp}}^{Q}$ are sampled according to the sampling ratio $\rho$. The state-value and action-value networks are optimized using the merged batch $\mathcal{B}_{\mathrm{mix}}$, whereas the policy network is updated using a separately sampled expert batch $\mathcal{B}_{\mathrm{exp}}^{\pi}$.

\subsection{Discussion}

The proposed framework can be interpreted as leveraging failure data to provide implicit negative feedback. Transitions leading to collisions are associated with low rewards, resulting in lower value estimates for unsafe state-action pairs. During policy extraction, these low-value actions are naturally avoided.

Unlike directly mixing failure data into policy learning, the proposed asymmetric design avoids introducing undesirable behaviors into the policy while still benefiting from the information contained in failure experiences.

%% file: Chapters/Experiment.tex
\section{Experimental Evaluation}

This section presents a comprehensive evaluation of the proposed framework in both simulation and real-world environments. The objective is to investigate whether the failure-aware learning framework can effectively improve navigation safety without degrading task performance. Specifically, we aim to answer the following questions:

\begin{itemize}
    \item Can failure data improve navigation safety when properly utilized?
    \item How does the proposed method compare with naive data mixing and demonstration-only learning?
    \item Does the learned policy generalize effectively from simulation to real-world environments and across different robot platforms?
\end{itemize}

\subsection{Simulation Evaluation}
\subsubsection{Dataset Acquisition}

\begin{table}[t]
    \centering
    \caption{List of compared methods.}
    \label{tab:description_methods}
    \renewcommand{\arraystretch}{1.15}
    \setlength{\tabcolsep}{6pt}
    \begin{tabular*}{\columnwidth}{@{\extracolsep{\fill}}lp{0.72\columnwidth}}
        \hline\hline
        Method & Description \\
        \hline
        IQL--CA & The proposed method trained on the structured mixed dataset. \\
        BC & Behavior cloning trained on the successful-only dataset. \\
        IQL--DM & Original IQL trained on the directly mixed dataset. \\
        IQL--SO & Original IQL trained on the successful-only dataset. \\
        \hline\hline
    \end{tabular*}
\end{table}

\begin{table}[t]
    \centering
    \caption{IQL-specific hyperparameters.}
    \label{tab:iql_hyper}
    \small
    \renewcommand{\arraystretch}{1.15}
    \setlength{\tabcolsep}{6pt}
    \begin{tabular*}{0.78\columnwidth}{@{\extracolsep{\fill}}ll}
        \hline\hline
        Parameter & Value \\
        \hline
        Expectile $\tau$ & 0.7 \\
        Discount factor $\gamma$ & 0.99 \\
        Target update coefficient $\alpha$ & 0.005 \\
        Weight temperature & 1.0 \\
        Max weight & 100 \\
        Number of critics & 2 \\
        \hline\hline
    \end{tabular*}
\end{table}

\begin{table*}[!t]
\centering
\caption{Simulation-based performance comparison of different training methods.}
\label{tab:simulation_result}
\renewcommand{\arraystretch}{1.10}
\resizebox{\textwidth}{!}{%
\setlength{\tabcolsep}{2.5pt}
\begin{tabular}{l llll llll llll}
\hline\hline
\multirow{2}{*}{Method}
& \multicolumn{4}{c}{Success Rate (\%)}
& \multicolumn{4}{c}{Collision Rate (\%)}
& \multicolumn{4}{c}{Timeout Rate (\%)} \\
\cmidrule(lr){2-5}\cmidrule(lr){6-9}\cmidrule(lr){10-13}
& SEnv1 & SEnv2 & SEnv3 & Overall
& SEnv1 & SEnv2 & SEnv3 & Overall
& SEnv1 & SEnv2 & SEnv3 & Overall \\
\hline
BC
& 93.33 $\pm$ 1.89 & 84.67 $\pm$ 3.77 & 68.67 $\pm$ 6.18 & 82.22 $\pm$ 3.95
& 5.33 $\pm$ 1.89 & 3.33 $\pm$ 2.49 & 30.67 $\pm$ 6.60 & 13.11 $\pm$ 3.66
& \textbf{1.33 $\pm$ 1.89} & 12.00 $\pm$ 1.63 & 0.67 $\pm$ 0.94 & 4.67 $\pm$ 1.49 \\
IQL--DM
& 90.80 $\pm$ 3.49 & 84.40 $\pm$ 3.44 & 67.20 $\pm$ 4.83 & 80.80 $\pm$ 3.92
& 7.20 $\pm$ 2.71 & 4.80 $\pm$ 2.04 & 32.40 $\pm$ 4.27 & 14.80 $\pm$ 3.01
& 2.00 $\pm$ 1.26 & 10.80 $\pm$ 2.04 & \textbf{0.40 $\pm$ 0.80} & 4.40 $\pm$ 1.37 \\
IQL--SO
& 92.60 $\pm$ 3.20 & 85.60 $\pm$ 3.67 & 71.60 $\pm$ 1.96 & 83.27 $\pm$ 2.94
& 6.00 $\pm$ 3.41 & 3.60 $\pm$ 2.06 & 27.40 $\pm$ 2.65 & 12.33 $\pm$ 2.71
& 1.40 $\pm$ 0.49 & 10.80 $\pm$ 3.19 & 1.00 $\pm$ 0.89 & 4.40 $\pm$ 1.52 \\
\rowcolor{gray!20}
IQL--CA
& \textbf{94.00 $\pm$ 1.63} & \textbf{88.00 $\pm$ 4.32} & \textbf{74.00 $\pm$ 1.63} & \textbf{85.33 $\pm$ 2.53}
& \textbf{4.67 $\pm$ 0.94} & \textbf{2.67 $\pm$ 1.89} & \textbf{25.33 $\pm$ 1.89} & \textbf{10.89 $\pm$ 1.57}
& \textbf{1.33 $\pm$ 0.94} & \textbf{9.33 $\pm$ 2.49} & 0.67 $\pm$ 0.94 & \textbf{3.78 $\pm$ 1.46} \\
\hline\hline
\end{tabular}}
\end{table*}

Data were collected in the lightweight \textit{ROS Stage} simulator~\cite{vaughan2008massively}, which provides high-fidelity 2--D simulation with low computational overhead. 
As shown in Fig.~\ref{fig:SEnvs}(\subref{fig:SEnv1}), data were collected in SEnv1, a $10\times10$~m$^2$ room with a cluttered layout. 
The data were collected using a policy trained by IPAPRec, which demonstrated strong navigation performance~\cite{zhang2022ipaprec}. During the data collection process, transition tuples from episodes that successfully reached the goal were stored in the success dataset $\mathcal{D}_{\mathrm{exp}}$, whereas those from episodes that ended in collision were stored in the failure dataset $\mathcal{D}_{\mathrm{col}}$. 
The control frequency was set to 5~Hz, and the maximum number of time steps per episode was $T_{\max}=200$. 

In the simulator, the robot was modeled as a differential-drive platform with a radius of 0.2~m, a maximum linear velocity of 0.5~m/s, and a maximum angular velocity of $\frac{\pi}{2}$~rad/s. 
It was equipped with a 2--D LiDAR sensor with a field of view of $270^\circ$, an angular resolution of $0.25^\circ$, and a maximum sensing range of 30~m. This dataset construction explicitly separates successful and collision transitions, which is essential for the proposed failure-aware learning framework.

\subsubsection{Model Training}

The model was trained in the offline RL setting using IQL as the backbone algorithm. 
The implementation was based on D3RLPY~\cite{d3rlpy}, and training was conducted on an NVIDIA GeForce RTX 4090 GPU. 

The training dataset contains 200k samples with a success-to-collision ratio of 9:1. 
The collision sampling ratio for critic network updates was set to 0.015, to ensure that failure data provide sufficient safety signals without dominating the training process and inducing excessively conservative behavior.

Both the actor and critic networks were optimized with Adam using learning rates of $3\times10^{-4}$, and the batch size was set to 256. Training was conducted for 300000 steps with 1000 steps per epoch, resulting in 300 epochs. Each method was trained with five random seeds, and model checkpoints were saved every 10 epochs. The IQL-specific hyperparameters are summarized in Table~\ref{tab:iql_hyper}.

\subsubsection{Simulation Results}

\begin{figure}[t]
    \centering
    \begin{subfigure}[b]{0.3\linewidth}
        \centering
        \includegraphics[width=\textwidth]{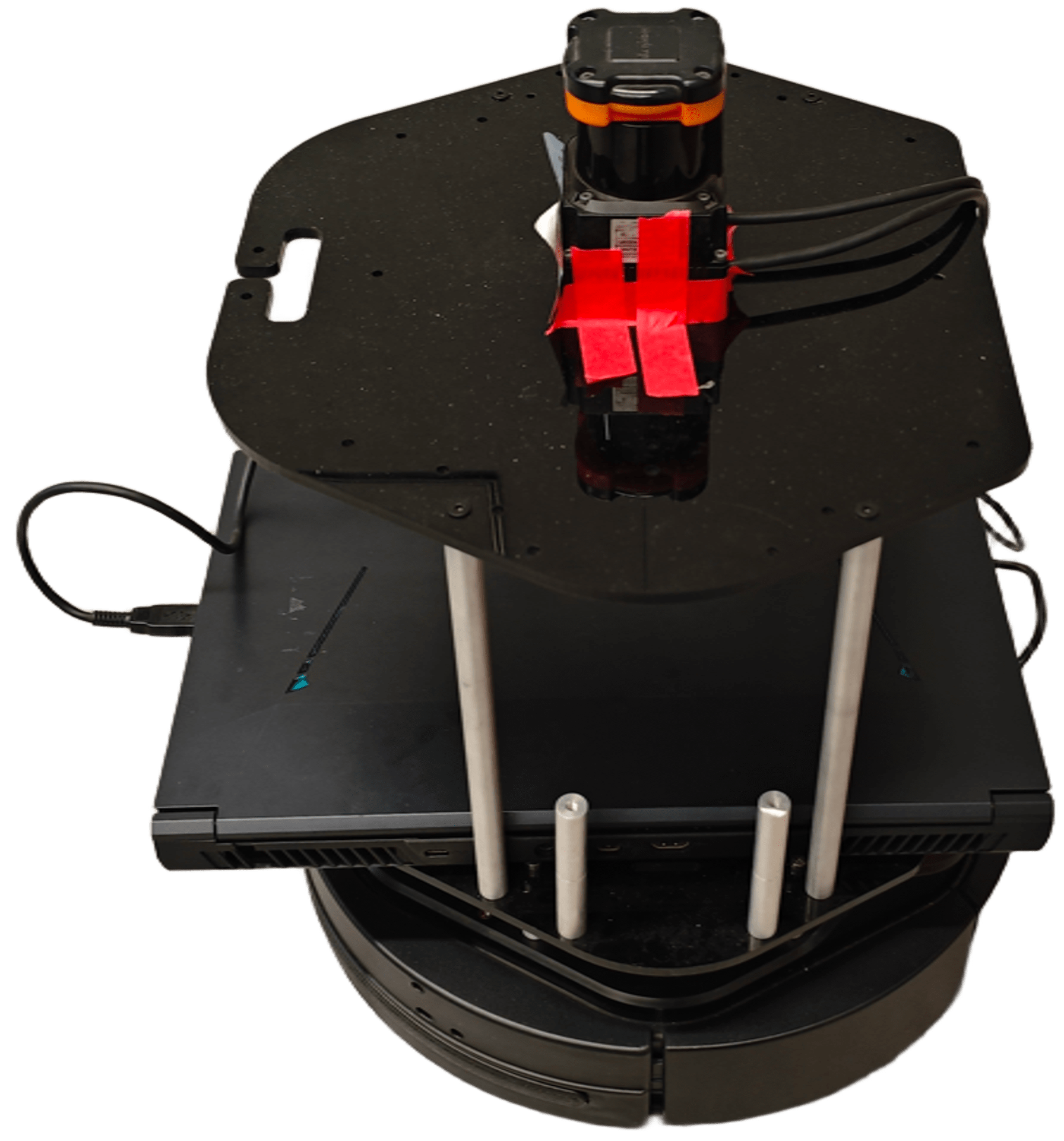}
        \caption{}
        \label{fig:turtlebot2}
    \end{subfigure}%
    \hspace{0.1\linewidth}%
    \begin{subfigure}[b]{0.3\linewidth}
        \centering
        \includegraphics[width=\textwidth]{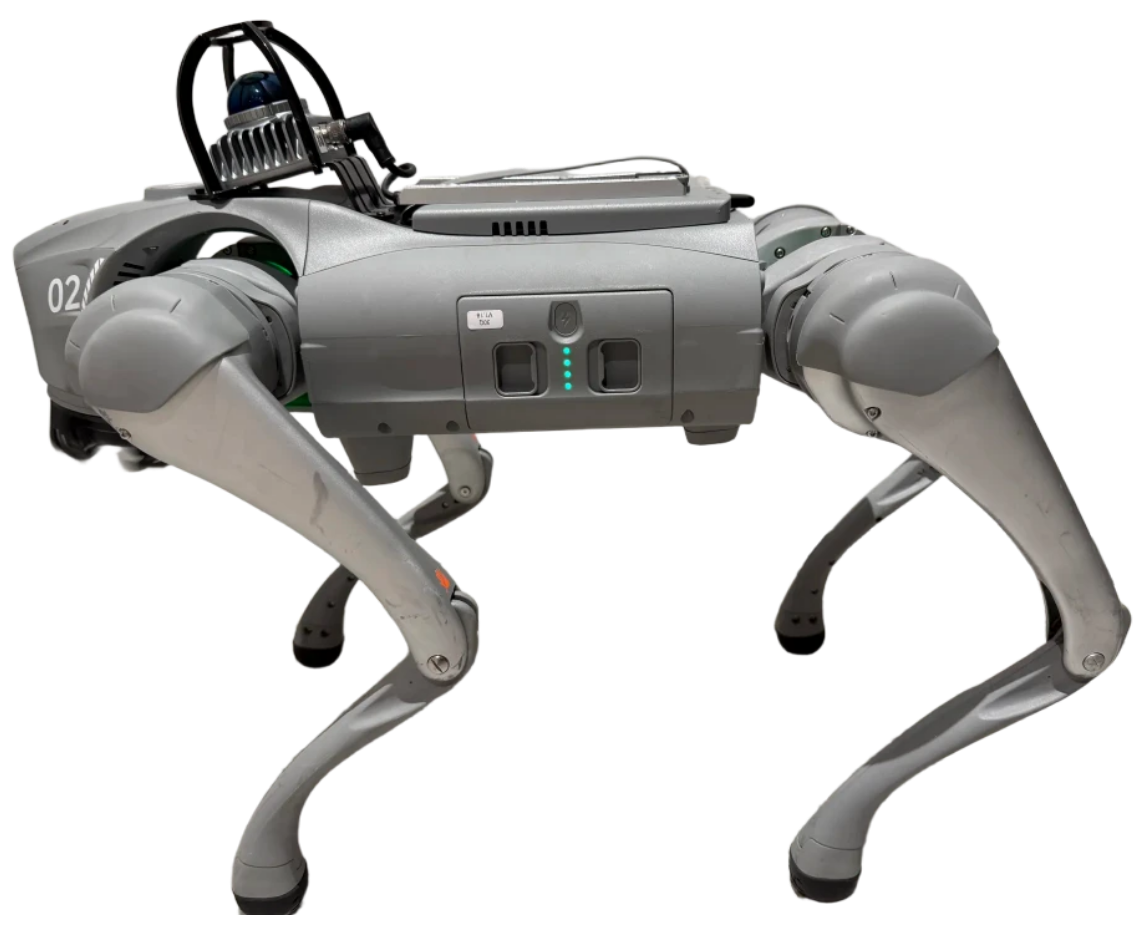}
        \caption{}
        \label{fig:unitree_go2}
    \end{subfigure}
    \caption{Robots for real-world experiments.}
    \label{fig:robot_figs}
\end{figure}

\begin{figure}[t]
    \centering
    \begin{subfigure}[b]{0.45\columnwidth}
        \centering
        \includegraphics[width=\textwidth]{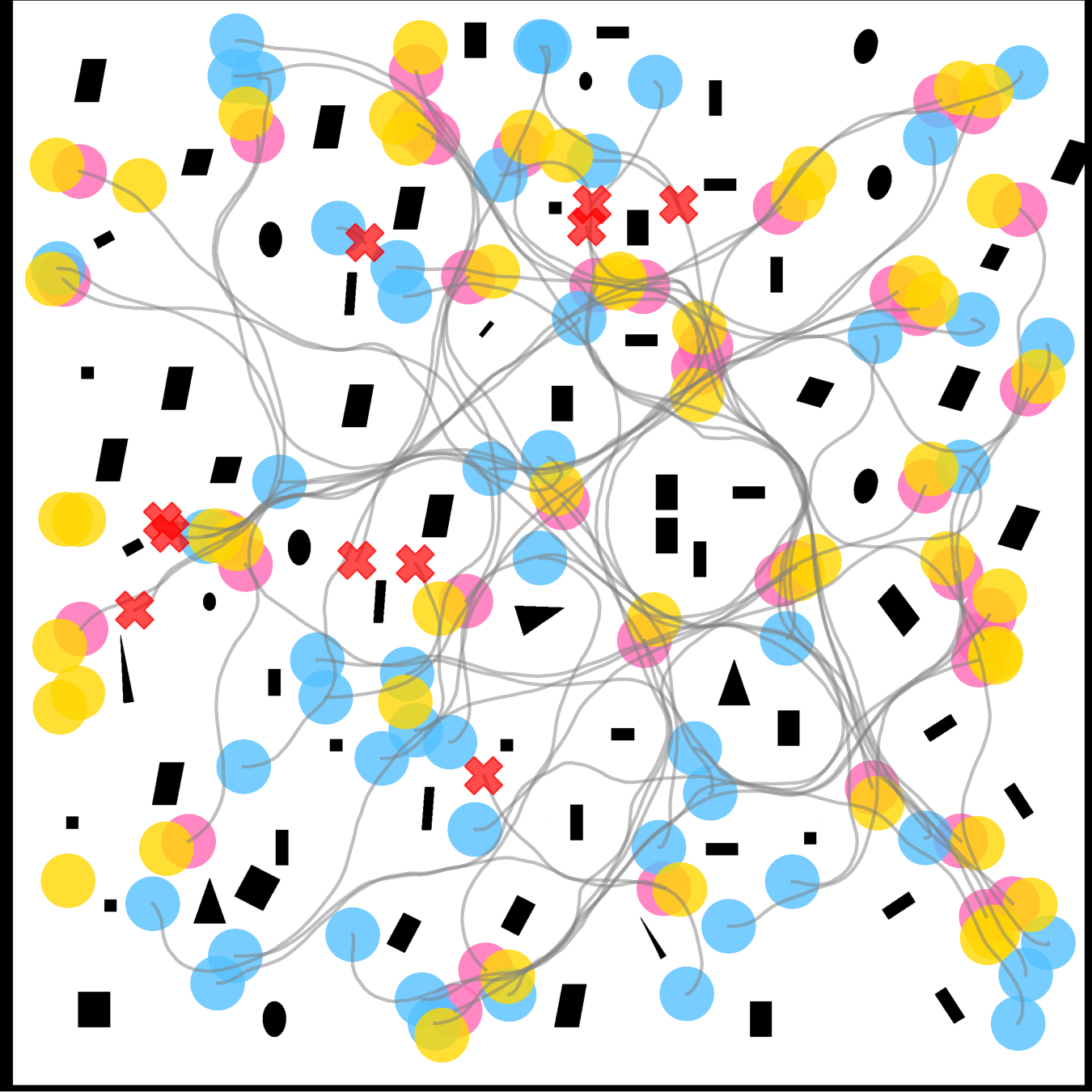}
        \caption{IQL--CA in SEnv3.}
        \label{fig:iql_ca_traj_SEnv3}
    \end{subfigure}
    \hfill
    \begin{subfigure}[b]{0.45\columnwidth}
        \centering
        \includegraphics[width=\textwidth]{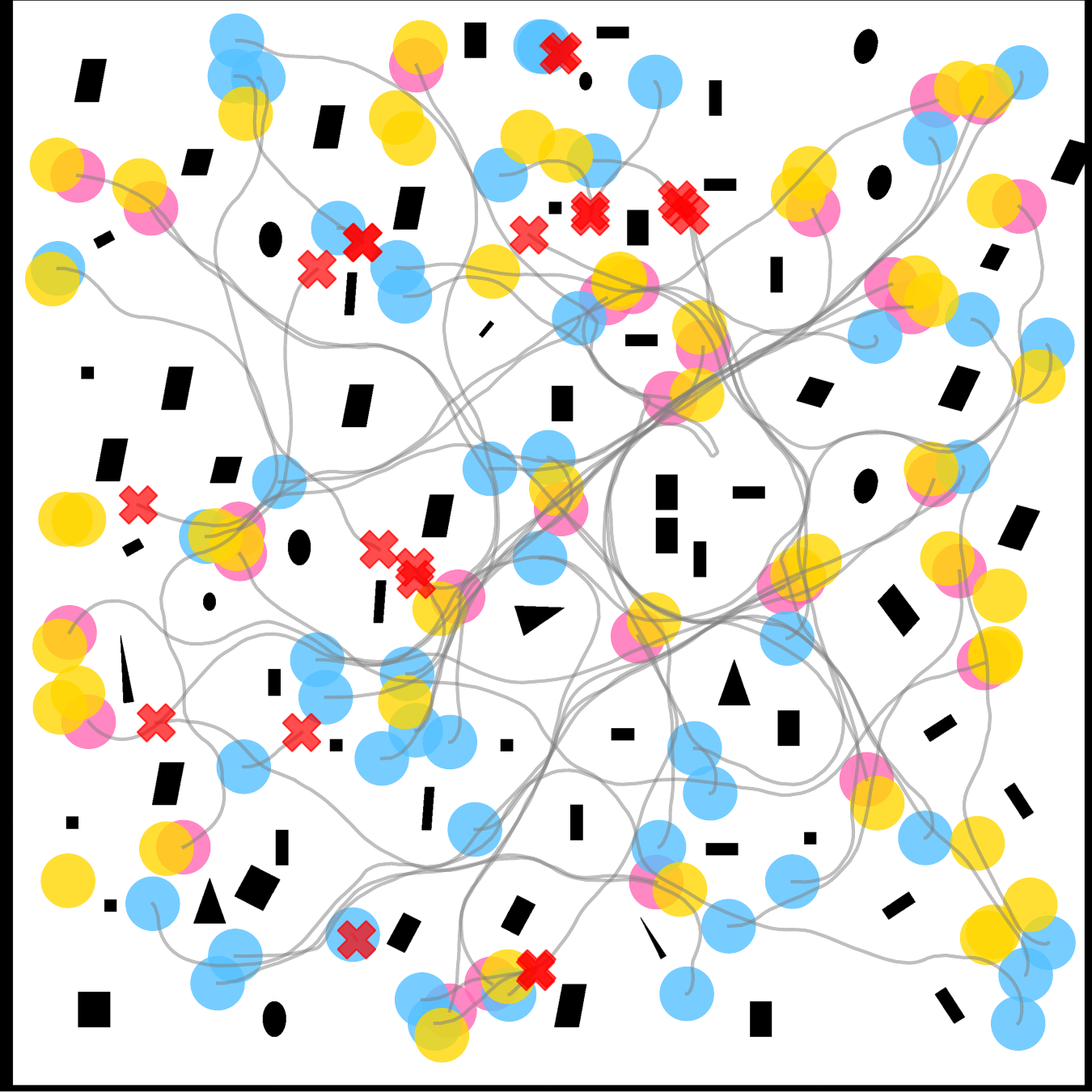}
        \caption{BC in SEnv3.}
        \label{fig:bc_traj_SEnv3}
    \end{subfigure}
    \hfill
    \begin{subfigure}[b]{0.45\columnwidth}
        \centering
        \includegraphics[width=\textwidth]{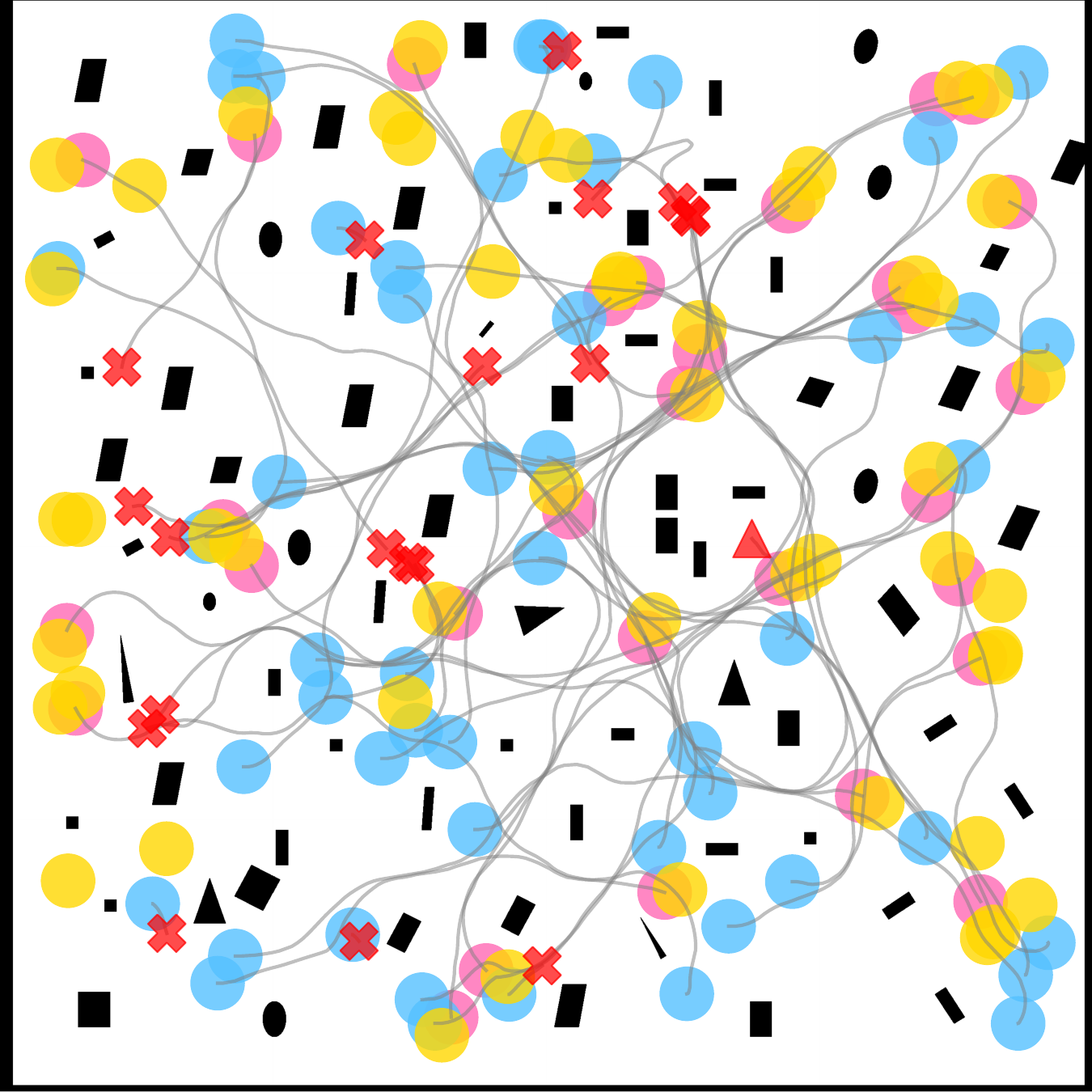}
        \caption{IQL--DM in SEnv3.}
        \label{fig:iql_dm_traj_SEnv3}
    \end{subfigure}
    \hfill
    \begin{subfigure}[b]{0.45\columnwidth}
        \centering
        \includegraphics[width=\textwidth]{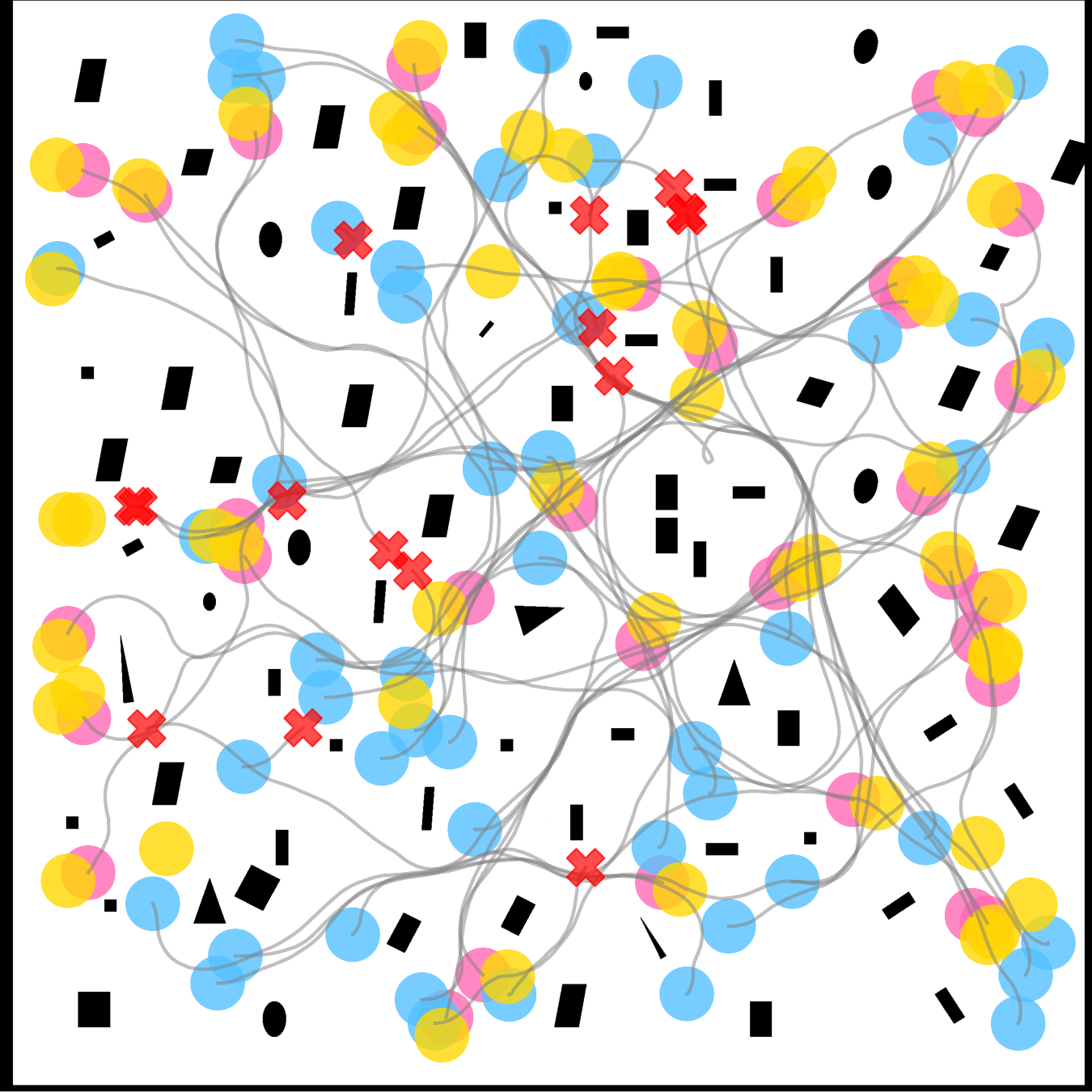}
        \caption{IQL--SO in SEnv3.}
        \label{fig:iql_so_traj_SEnv3}
    \end{subfigure}
    \caption{Trajectories of robots in SEnv3. Start, goal, success, crash, and timeout states are marked by blue circle, yellow circle, pink circle, red cross, and red triangle symbols, respectively.}
    \label{fig:traj_in_SEnv3}
\end{figure}

This section presents a performance evaluation of the proposed method in the three simulated scenarios shown in Fig.~\ref{fig:SEnvs}. The evaluation was performed in two $10\times10$~m$^2$ rooms, shown in Fig.~\ref{fig:SEnvs}(\subref{fig:SEnv1}) and Fig.~\ref{fig:SEnvs}(\subref{fig:SEnv3}), and one $8\times8$~m$^2$ room, shown in Fig.~\ref{fig:SEnvs}(\subref{fig:SEnv2}). These scenarios feature different obstacle densities to emulate real office environments.

During evaluation, 50 point-to-point navigation tasks were conducted in each scenario across three independent trials. In each task, the robot was required to move from a designated start position to a goal location within the maximum allowed number of steps. The objective was to complete these tasks with a low collision rate while maintaining a high success rate.

We use three performance metrics: average success rate (SR), average collision rate (CR), and average timeout rate (TR). The compared methods are summarized in Table~\ref{tab:description_methods}, and the quantitative results are reported in Table~\ref{tab:simulation_result}.

Overall, IQL--CA achieves the highest success rate while consistently reducing the collision rate across all testing scenarios. Compared with BC, which is trained only on successful demonstrations, IQL--CA attains a notable reduction in the collision rate, especially in the most challenging scenario, SEnv3. Specifically, the collision rate is reduced by 5.34\% compared with BC, indicating that failure experiences provide useful safety information beyond successful demonstrations.

Furthermore, IQL--CA yields a lower collision rate than IQL--SO, illustrating that offline RL trained solely on successful data is insufficient to achieve equivalent safety benefits. More importantly, IQL--DM, which learns from a dataset obtained by directly mixing success and collision data, exhibits the worst performance in both success rate and collision rate. This result suggests that the performance gain does not stem merely from the inclusion of collision data, but rather from utilizing such data under the failure-aware learning principle.

\subsubsection{Effect of Failure Data}
We further analyze the impact of failure data through a comparison of three settings: (1) IQL--SO, (2) IQL--DM, and (3) IQL--CA, as detailed in Table~\ref{tab:description_methods}. 

The results show that failure data alone are not sufficient to improve performance when used without structure. 
Naive mixing leads to degraded performance, whereas the proposed method achieves the best trade-off between safety and navigation success. This confirms that the effectiveness of failure data depends not only on their presence, but also on how they are incorporated into learning.

\subsubsection{Qualitative Analysis}

To further illustrate the qualitative difference, Fig.~\ref{fig:traj_in_SEnv3} visualizes the robot trajectories in SEnv3. 
The trajectories generated by IQL--CA controller exhibit smoother and safer navigation behavior, with fewer collisions in cluttered regions. In contrast, BC and IQL--DM controllers frequently collide with obstacles in narrow passages, while IQL--SO controller remains less safe than IQL--CA controller. These results further support the effectiveness of the proposed failure-aware learning framework.

\subsection{Real-world Evaluation}

\begin{figure*}[!t]
    \centering
    \captionsetup[subfigure]{skip=2pt}

    \begin{subfigure}[b]{0.23\textwidth}
        \centering
        \includegraphics[width=\linewidth,trim=6 6 6 6,clip]{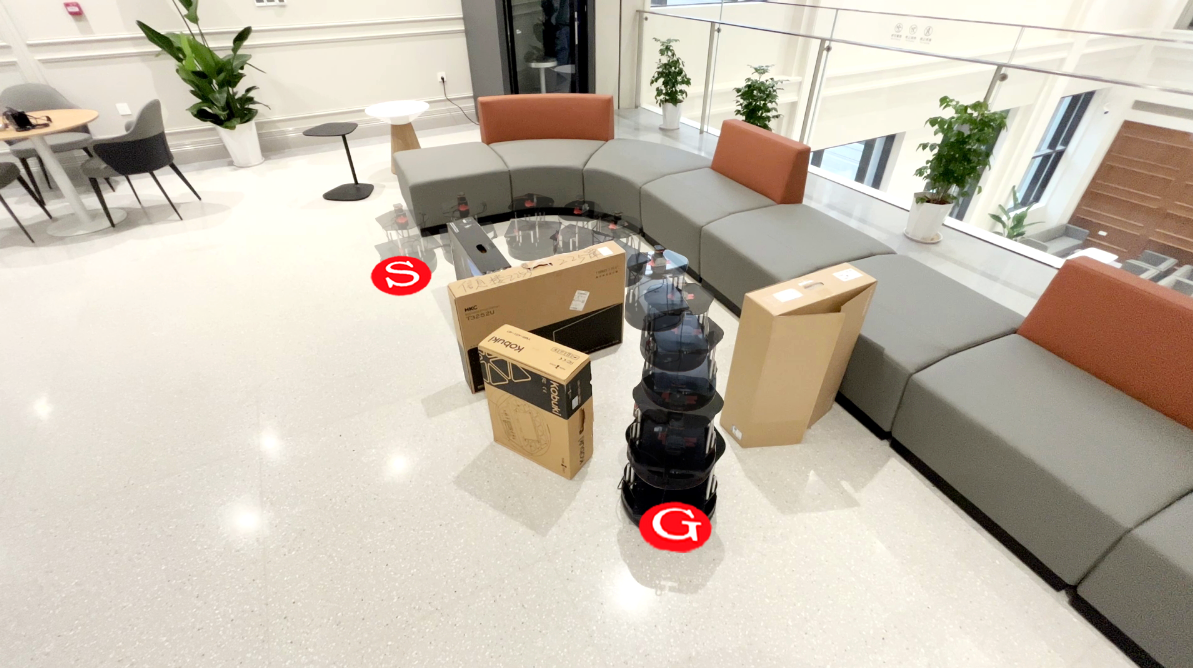}
        \caption{IQL--CA in REnv1.}
        \label{fig:IQL_CA_REnv1}
    \end{subfigure}\hspace{0.005\textwidth}%
    \begin{subfigure}[b]{0.23\textwidth}
        \centering
        \includegraphics[width=\linewidth,trim=6 6 6 6,clip]{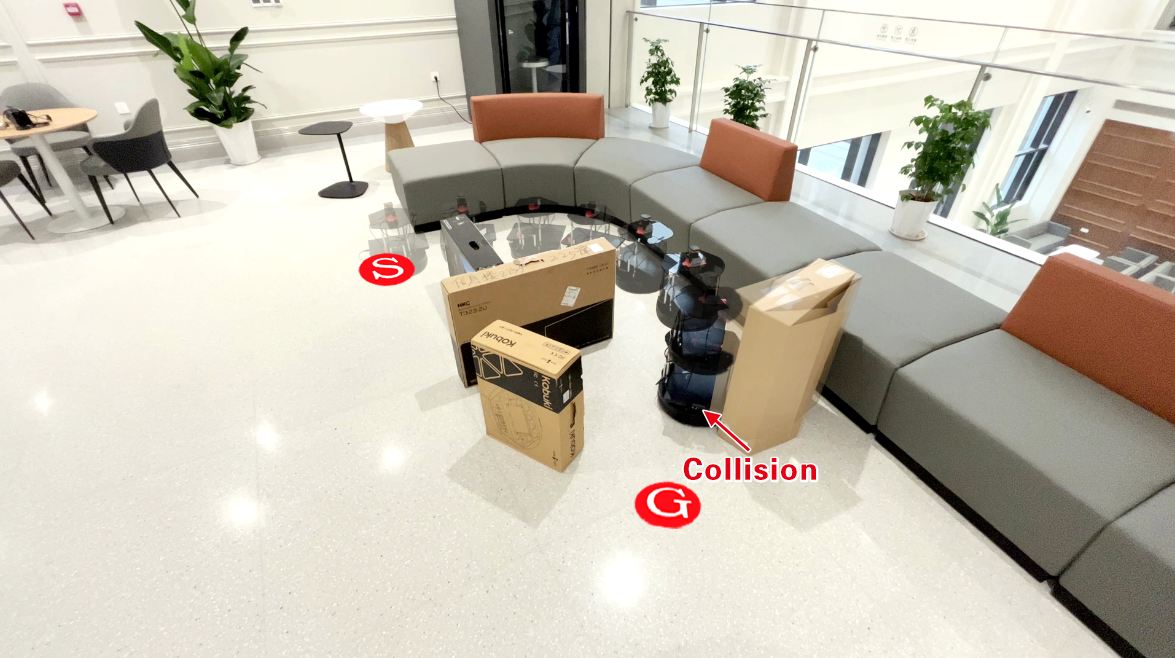}
        \caption{BC in REnv1.}
        \label{fig:BC_REnv1}
    \end{subfigure}\hspace{0.005\textwidth}%
    \begin{subfigure}[b]{0.23\textwidth}
        \centering
        \includegraphics[width=\linewidth,trim=6 6 6 6,clip]{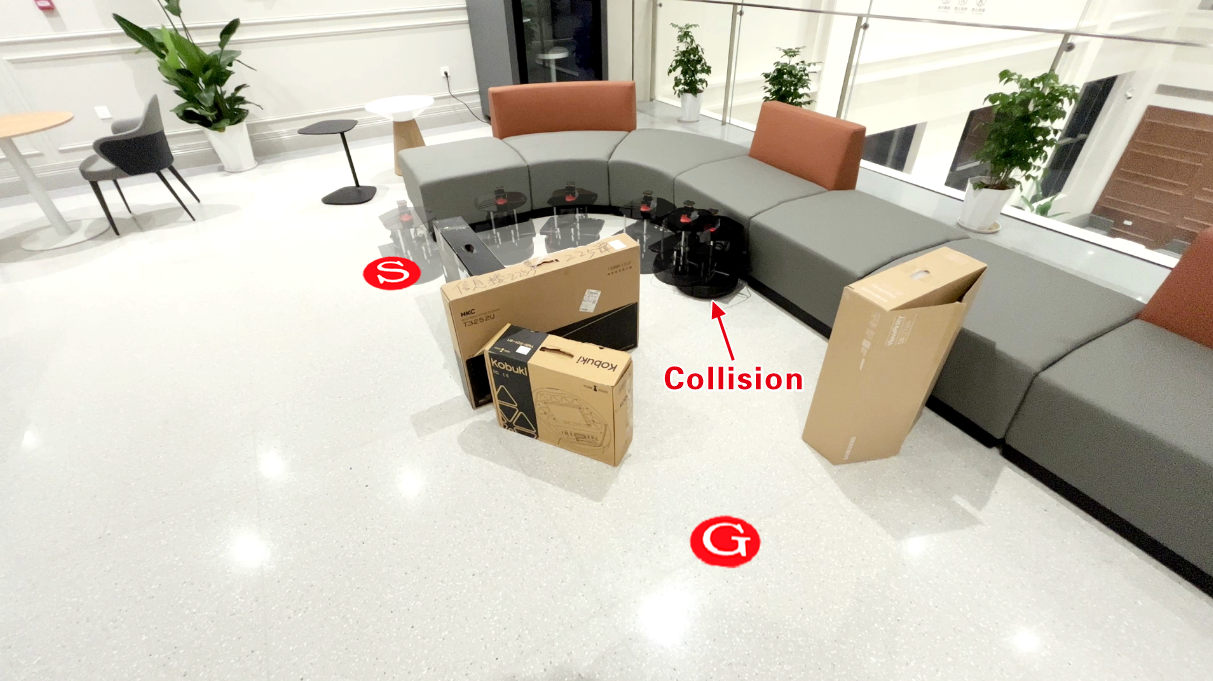}
        \caption{IQL--DM in REnv1.}
        \label{fig:IQL_DM_REnv1}
    \end{subfigure}\hspace{0.005\textwidth}%
    \begin{subfigure}[b]{0.23\textwidth}
        \centering
        \includegraphics[width=\linewidth,trim=6 6 6 6,clip]{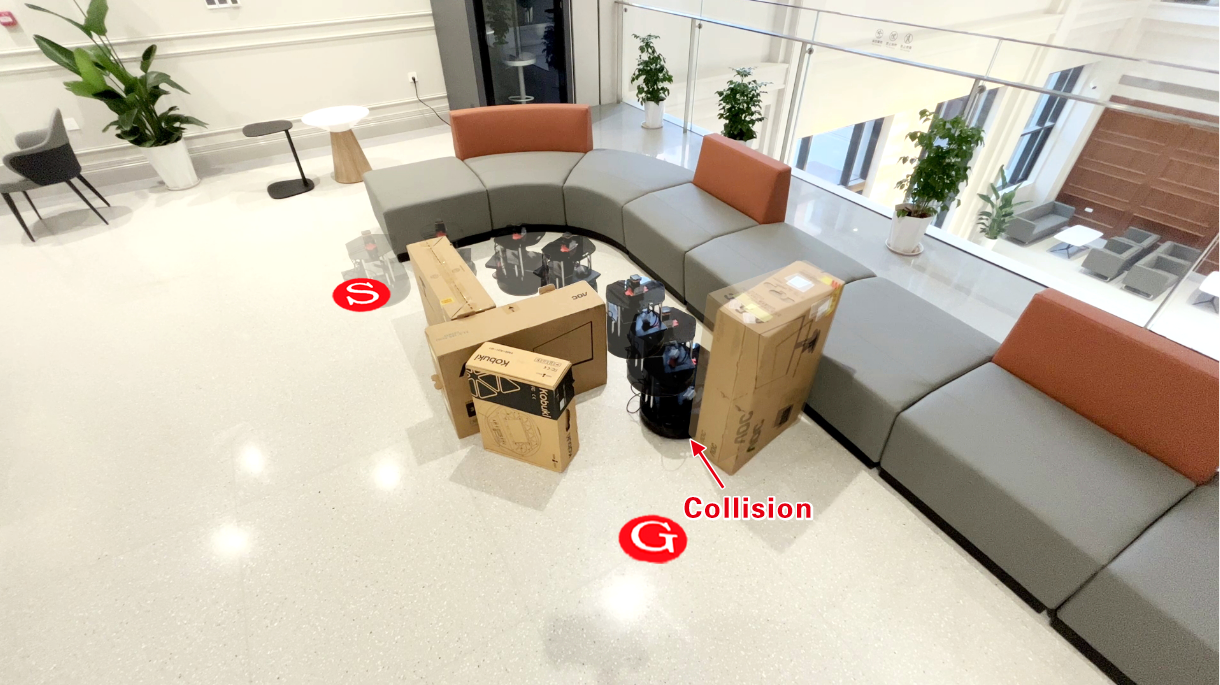}
        \caption{IQL--SO in REnv1.}
        \label{fig:IQL_SO_REnv1}
    \end{subfigure}

    \vspace{0.1em}

    \begin{subfigure}[b]{0.23\textwidth}
        \centering
        \includegraphics[width=\linewidth,trim=6 6 6 6,clip]{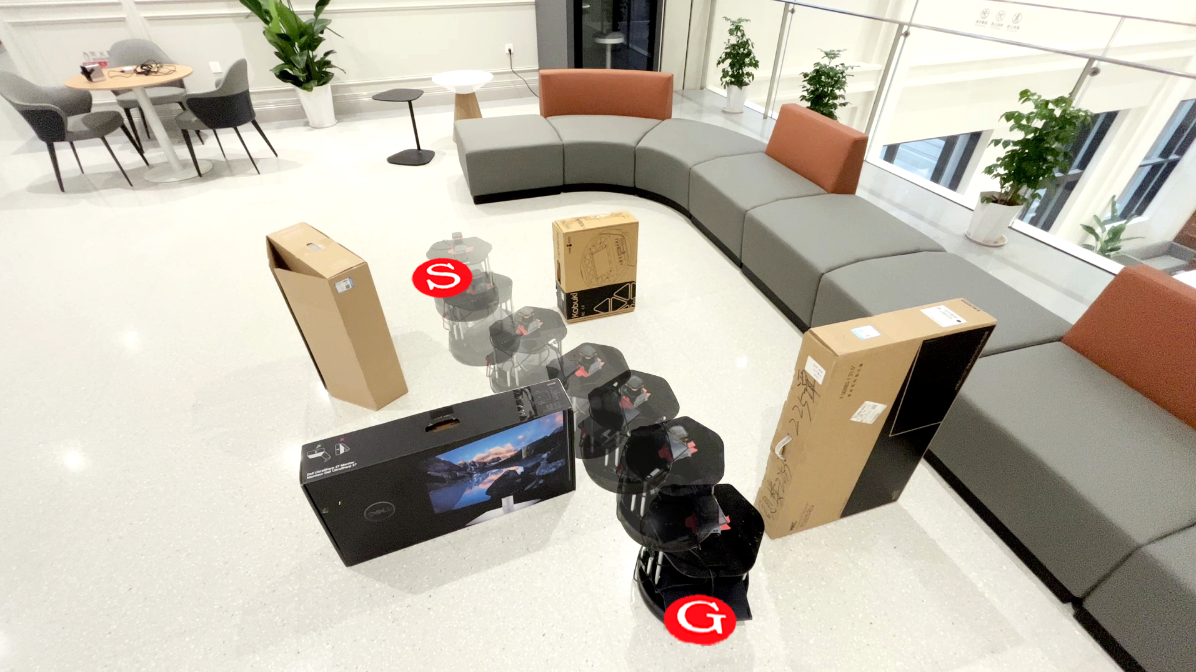}
        \caption{IQL--CA in REnv2.}
        \label{fig:IQL_CA_REnv2}
    \end{subfigure}\hspace{0.005\textwidth}%
    \begin{subfigure}[b]{0.23\textwidth}
        \centering
        \includegraphics[width=\linewidth,trim=6 6 6 6,clip]{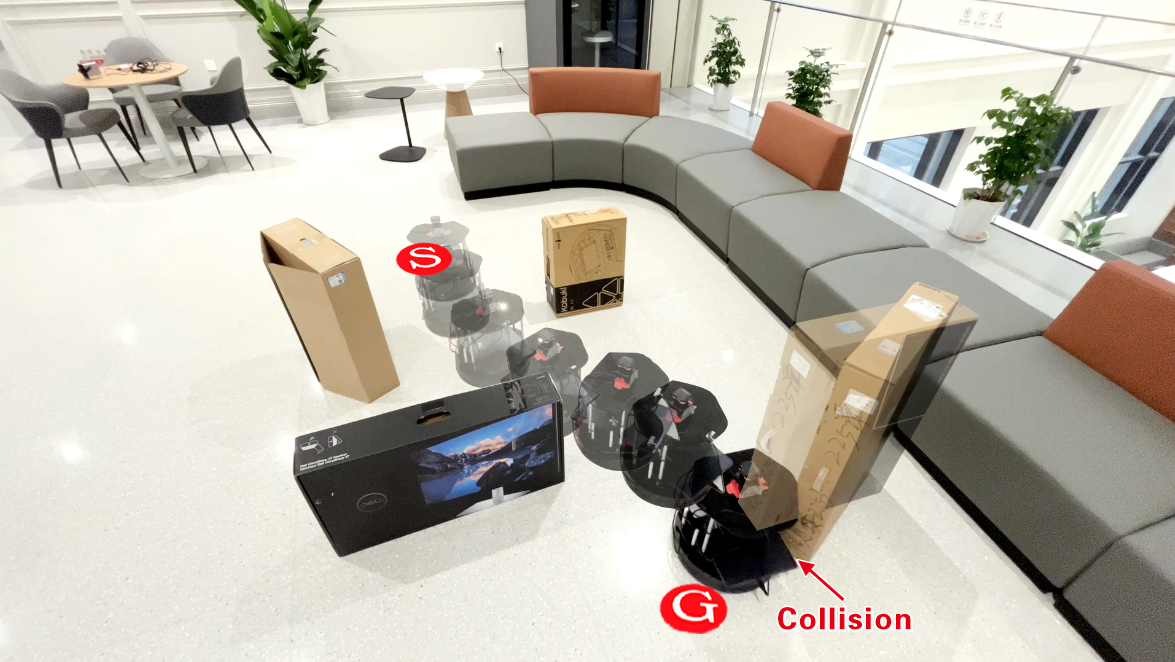}
        \caption{BC in REnv2.}
        \label{fig:BC_REnv2}
    \end{subfigure}\hspace{0.005\textwidth}%
    \begin{subfigure}[b]{0.23\textwidth}
        \centering
        \includegraphics[width=\linewidth,trim=6 6 6 6,clip]{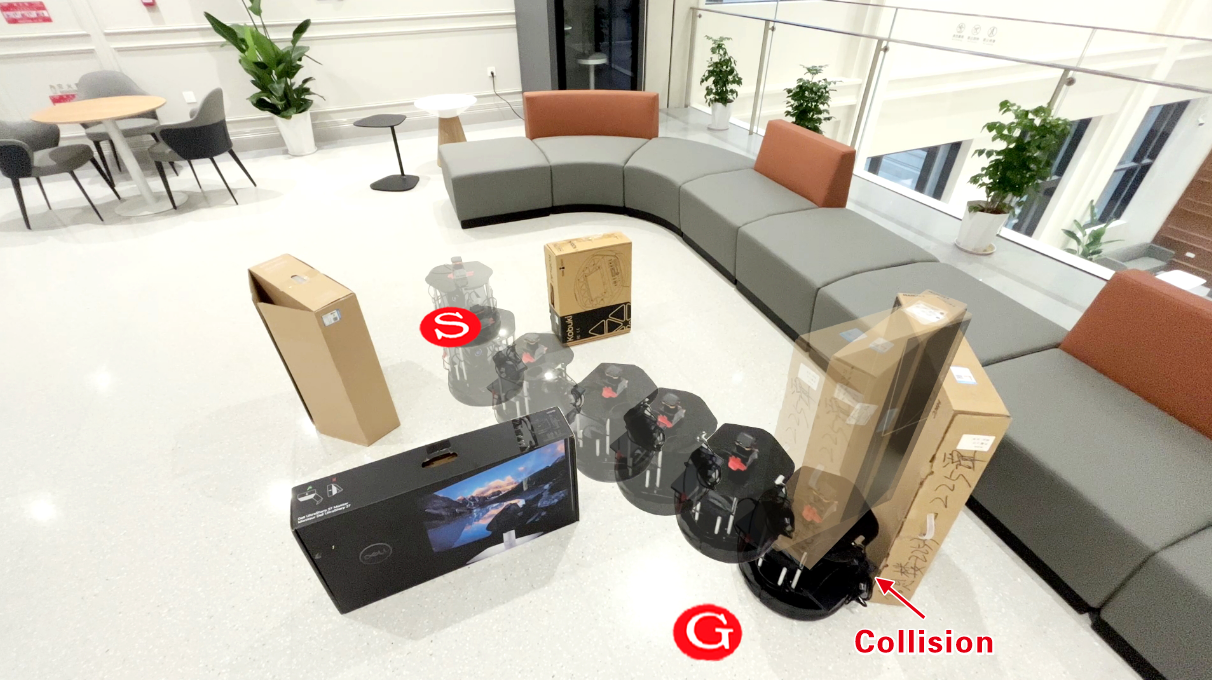}
        \caption{IQL--DM in REnv2.}
        \label{fig:IQL_DM_REnv2}
    \end{subfigure}\hspace{0.005\textwidth}%
    \begin{subfigure}[b]{0.23\textwidth}
        \centering
        \includegraphics[width=\linewidth,trim=6 6 6 6,clip]{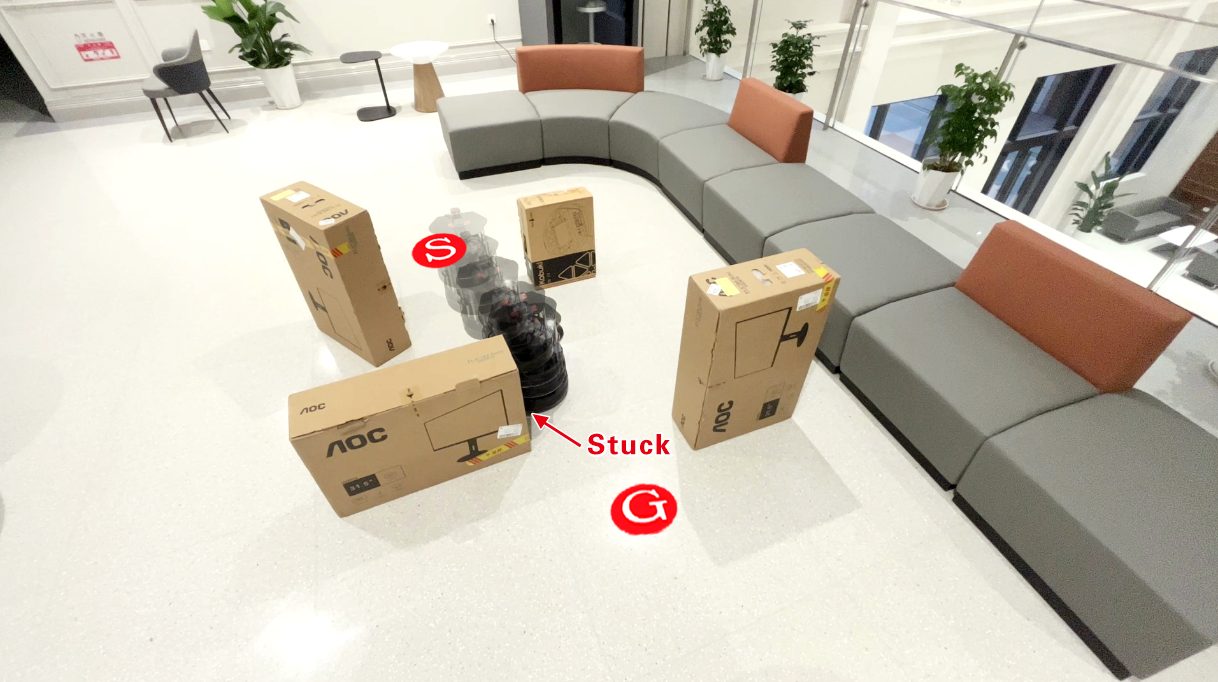}
        \caption{IQL--SO in REnv2.}
        \label{fig:IQL_SO_REnv2}
    \end{subfigure}

    \vspace{0.1em}

    \begin{subfigure}[b]{0.23\textwidth}
        \centering
        \includegraphics[width=\linewidth,trim=6 6 6 6,clip]{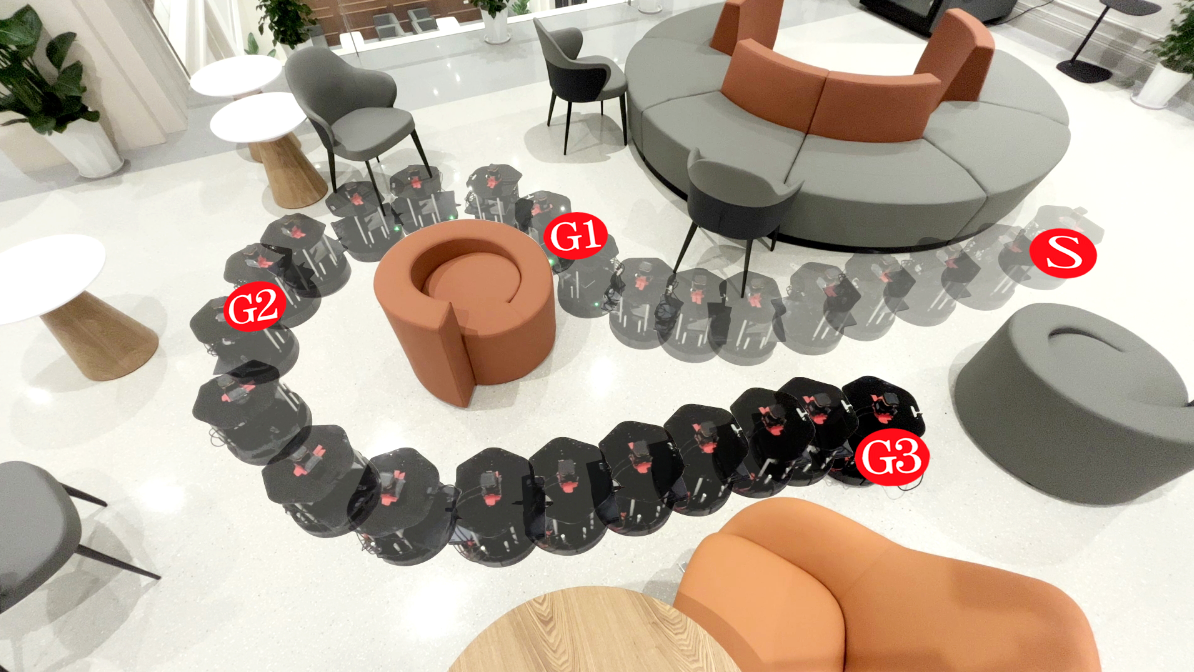}
        \caption{IQL--CA in REnv3.}
        \label{fig:IQL_CA_REnv3}
    \end{subfigure}\hspace{0.005\textwidth}%
    \begin{subfigure}[b]{0.23\textwidth}
        \centering
        \includegraphics[width=\linewidth,trim=6 6 6 6,clip]{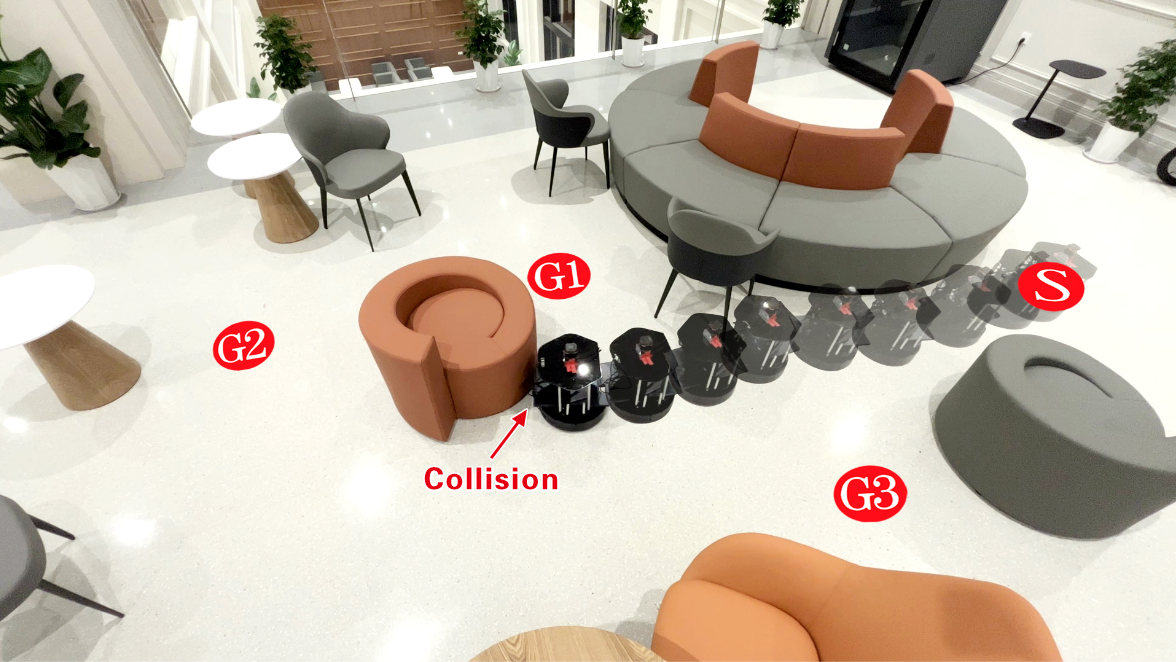}
        \caption{BC in REnv3.}
        \label{fig:BC_REnv3}
    \end{subfigure}\hspace{0.005\textwidth}%
    \begin{subfigure}[b]{0.23\textwidth}
        \centering
        \includegraphics[width=\linewidth,trim=6 6 6 6,clip]{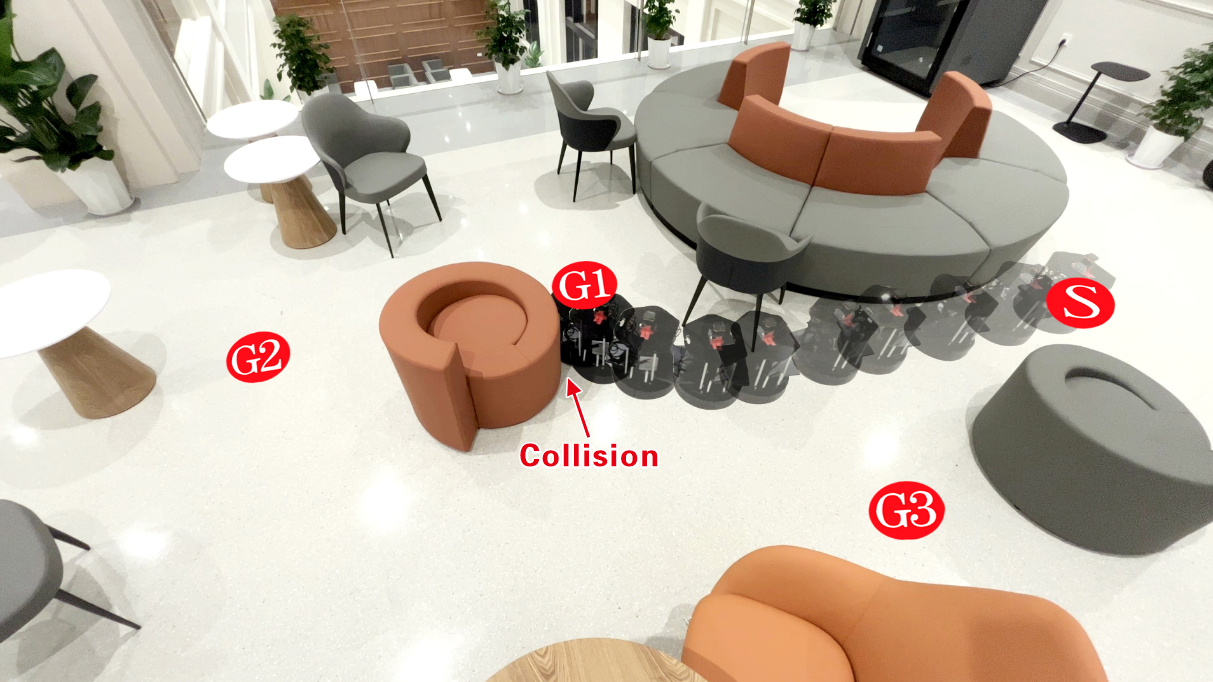}
        \caption{IQL--DM in REnv3.}
        \label{fig:IQL_DM_REnv3}
    \end{subfigure}\hspace{0.005\textwidth}%
    \begin{subfigure}[b]{0.23\textwidth}
        \centering
        \includegraphics[width=\linewidth,trim=6 6 6 6,clip]{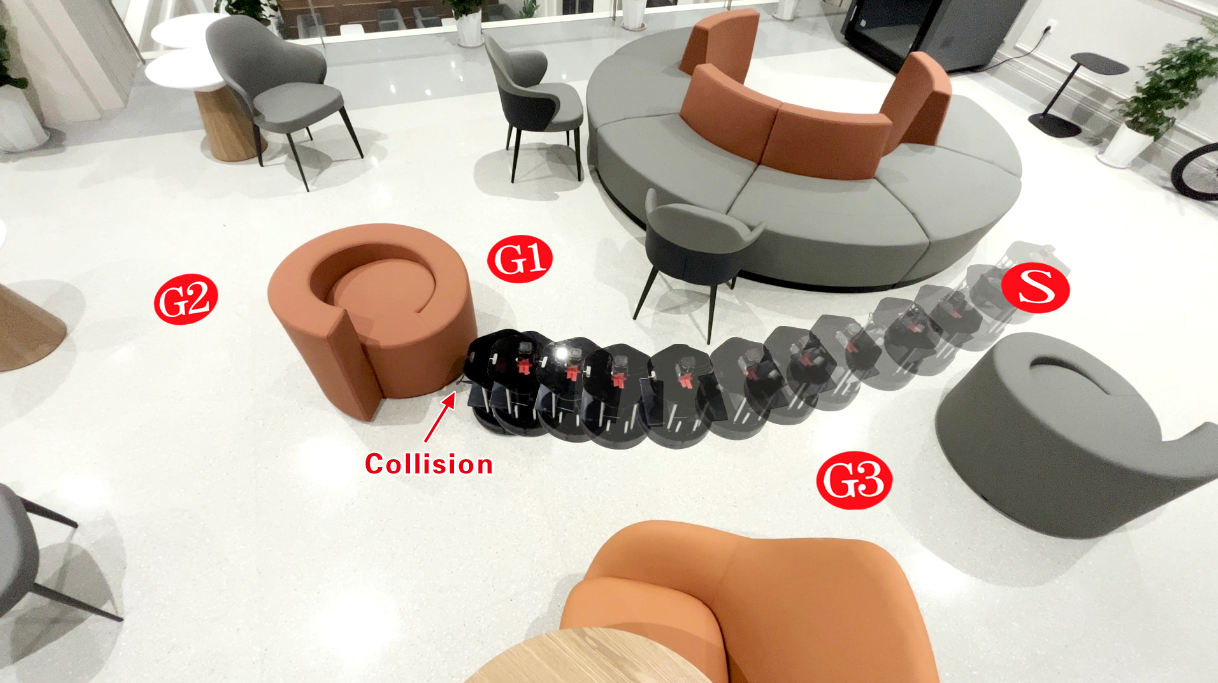}
        \caption{IQL--SO in REnv3.}
        \label{fig:IQL_SO_REnv3}
    \end{subfigure}

    \caption{Trajectories of the robot under different training methods when tested in REnv1--3. The experimental videos can be found in the supplementary file.}
    \label{fig:real_world_exp_turtlebot}
\end{figure*}

\begin{figure}[t]
    \centering
    \captionsetup[subfigure]{skip=2pt}

    \begin{subfigure}[b]{0.45\linewidth}
        \centering
        \includegraphics[width=\linewidth,trim=6 6 6 6,clip]{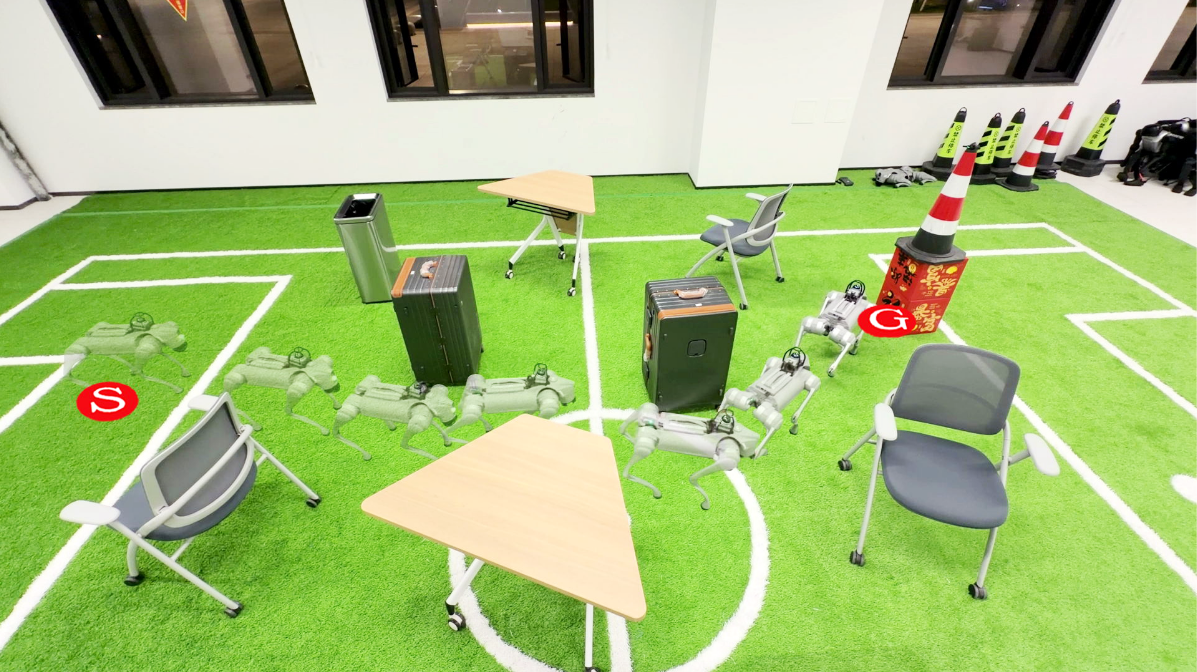}
        \caption{IQL--CA in REnv5.}
        \label{fig:IQL_CA_REnv5}
    \end{subfigure}\hspace{0.01\linewidth}%
    \begin{subfigure}[b]{0.45\linewidth}
        \centering
        \includegraphics[width=\linewidth,trim=6 6 6 6,clip]{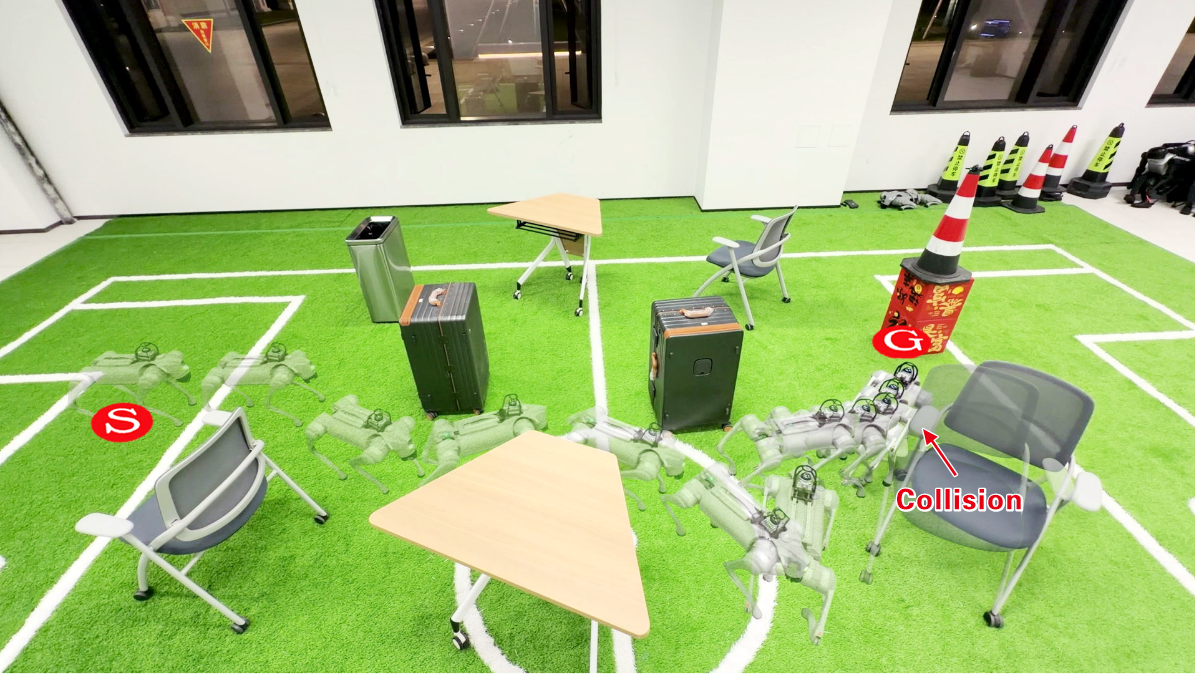}
        \caption{BC in REnv5.}
        \label{fig:BC_REnv5}
    \end{subfigure}

    \vspace{0.1em}

    \begin{subfigure}[b]{0.45\linewidth}
        \centering
        \includegraphics[width=\linewidth,trim=6 6 6 6,clip]{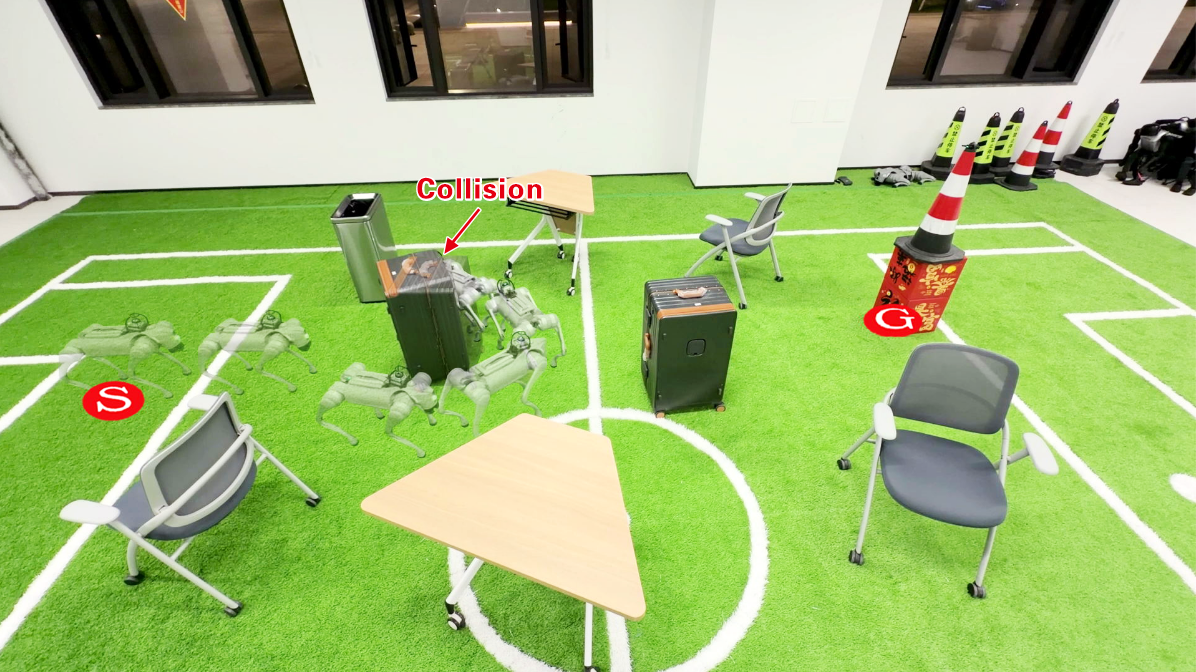}
        \caption{IQL--DM in REnv5.}
        \label{fig:IQL_DM_REnv5}
    \end{subfigure}\hspace{0.01\linewidth}%
    \begin{subfigure}[b]{0.45\linewidth}
        \centering
        \includegraphics[width=\linewidth,trim=6 6 6 6,clip]{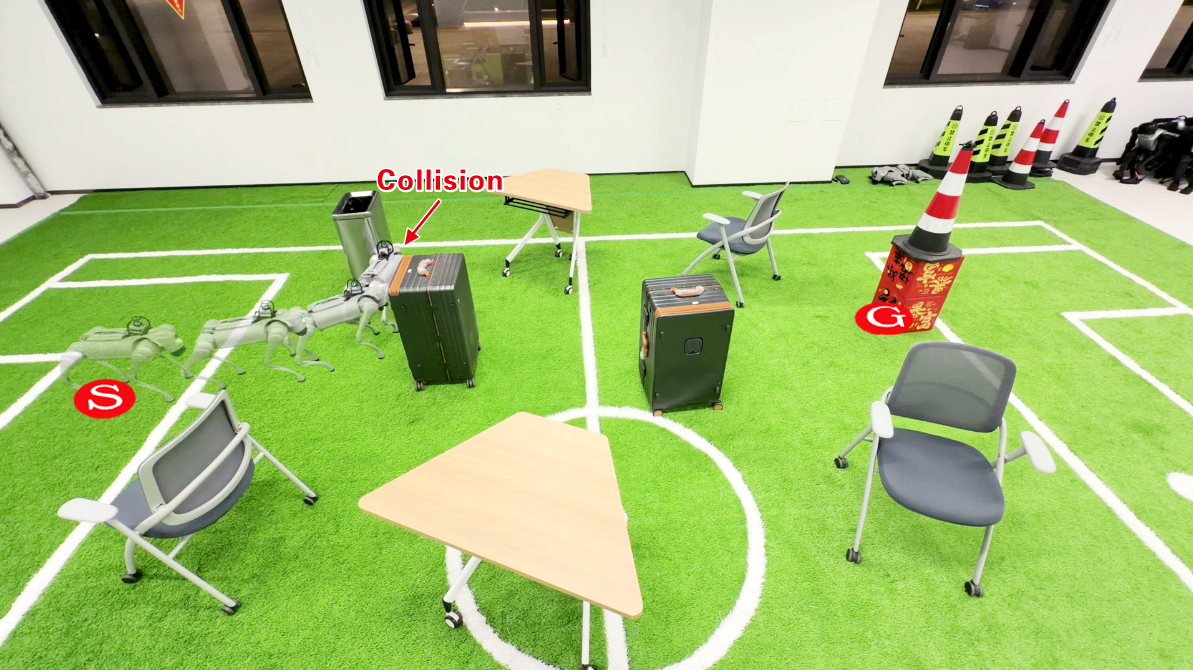}
        \caption{IQL--SO in REnv5.}
        \label{fig:IQL_SO_REnv5}
    \end{subfigure}

    \vspace{0.1em}

    \begin{subfigure}[b]{0.9\linewidth}
        \centering
        \includegraphics[width=\linewidth]{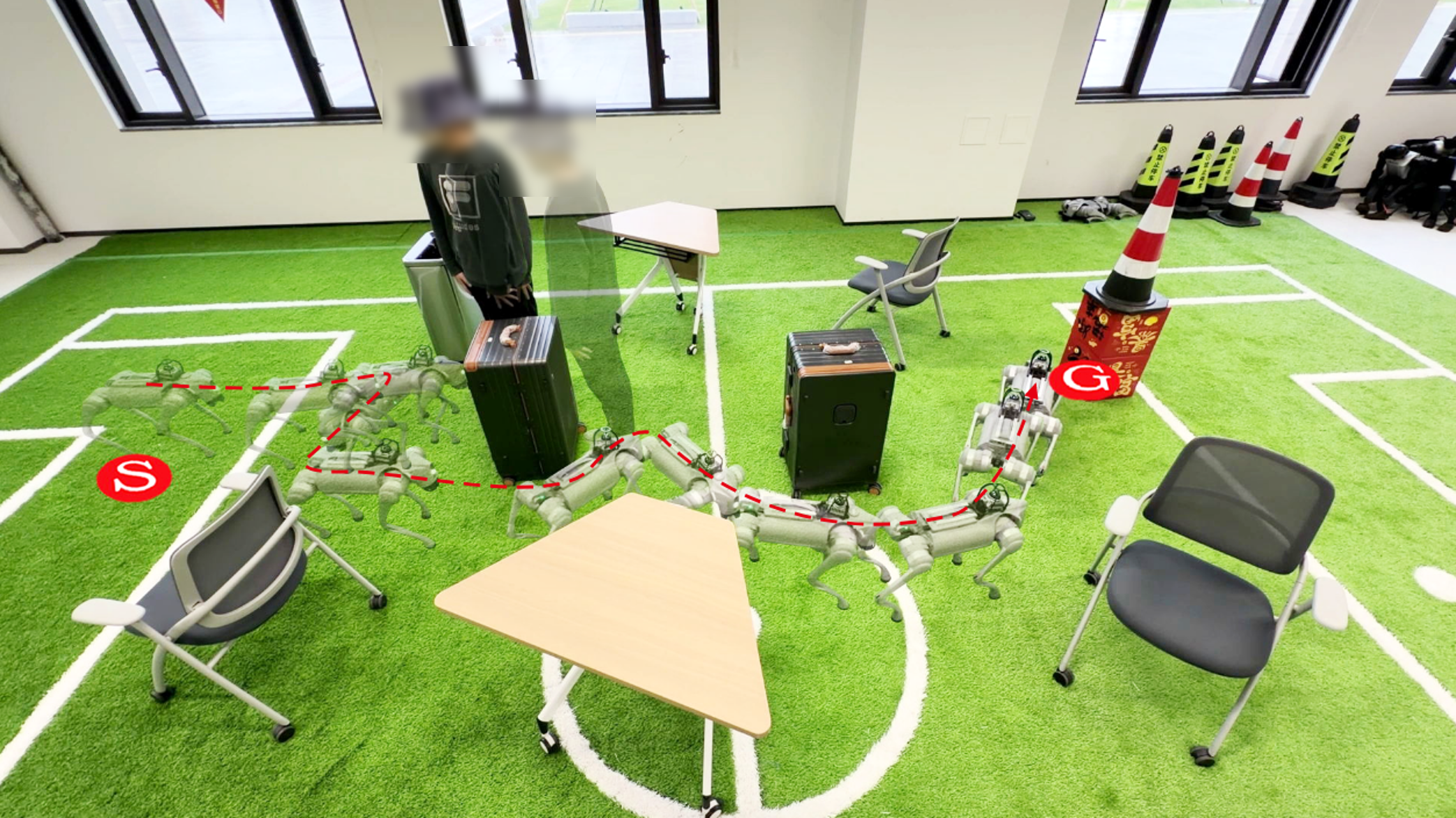}
        \caption{IQL--CA in REnv5 during the dynamic test.}
        \label{fig:IQL_CA_dynamic}
    \end{subfigure}

    \caption{Trajectories of the Unitree Go2 in REnv5 under different training configurations.}
    \label{fig:real_world_exp_go2}
\end{figure}

\subsubsection{Experimental Setup}

To further evaluate the navigation capabilities of the IQL--CA controller, we conducted extensive real-world experiments on different robot platforms under both static and dynamic scenarios. The navigation policy is trained exclusively in simulation, while all real-world testing scenarios remain unseen during training, enabling us to assess its generalization capability. Specifically, the evaluation is carried out on two representative platforms:
\begin{itemize}
    \item \textit{Mobile Robot Platform:} A Turtlebot2 robot (Fig. \ref{fig:turtlebot2}) equipped with a Hokuyo UTM-30LX 2--D LiDAR. The onboard controller was a PC with an Intel i9-12900H CPU, and no GPU was used. Target localization was achieved using a prebuilt map of the testing scenario generated by ROS GMapping~\cite{tai2017virtual,xie2020learning,Gmapping} and robot pose estimation from ROS AMCL~\cite{AMCL}. The map was used only to derive the target position in the robot frame from the robot pose and goal coordinates, rather than for motion planning.
    \item \textit{Legged Robot Platform:} A Unitree Go2 robot (Fig. \ref{fig:unitree_go2}) equipped with a Livox MID-360 3--D LiDAR. The onboard controller was an NVIDIA Jetson Orin NX, providing 100\,TOPS of AI computing power and operating independently without external computational support. Target localization was achieved using UWB.
\end{itemize}

\subsubsection{Testing Scenarios and Task Description}

For the mobile robot, three real-world scenarios with different travel distances and obstacle configurations were adopted. 
As shown in Fig.~\ref{fig:real_world_exp_turtlebot}, obstacles were manually arranged in REnv1 and REnv2 to emulate cluttered indoor environments with dense objects, while REnv3 was a natural indoor office environment used to validate navigation performance in a realistic workplace. 
For the legged robot, as illustrated in Fig.~\ref{fig:real_world_exp_go2}, an additional daily-life scenario was adopted to further evaluate whether the navigation capability learned on the mobile robot can generalize to a different robot platform and a more realistic scenario.

In each scenario, the robot was initialized at the start point `S' and was required to reach the goal point `G'. 
For scenarios with multiple goals, the robot was required to navigate successfully to each goal, denoted as `G$i$'. 
After reaching one goal, the robot would pause briefly to indicate successful arrival and then proceed to the next goal. 
In addition, for the legged robot navigation task, the robot was considered to have reached the goal when it was within 0.5~m of the goal position.

\subsubsection{Real-World Results}

For the mobile robot, the trajectories obtained in different real-world scenarios are shown in Fig.~\ref{fig:real_world_exp_turtlebot}. 
The experimental videos can be found in the supplementary file. 
As shown, the robot controlled by IQL--CA consistently demonstrated strong navigation performance under different obstacle densities and configurations. 
It successfully completed the navigation tasks without collisions while navigating through narrow spaces and around obstacles with high precision. 
In contrast, the other methods experienced collisions or became stuck in these scenarios.

Additionally, we used the legged robot to evaluate whether the navigation capability could generalize across different robot platforms. 
As shown in Fig.~\ref{fig:real_world_exp_go2}, only IQL--CA enabled the robot to successfully traverse the obstacles and reach the goal, whereas the other methods all resulted in collisions. These results further demonstrate that the proposed method generalizes beyond the mobile robot platform used during data collection.

\subsubsection{Dynamic Scenario Evaluation}

We further assessed the ability of IQL--CA controller to handle dynamic obstacles. 
Fig.~\ref{fig:real_world_exp_go2}(\subref{fig:IQL_CA_dynamic}) shows that the robot successfully avoided a pedestrian who appeared suddenly while continuing toward the goal. 
This result demonstrates that IQL--CA controller can effectively handle sudden pedestrian interactions during navigation, which are common in real-world environments.

\subsection{Summary}

Overall, results from both simulation and real-world experiments consistently demonstrate the effectiveness of our method. The proposed failure-aware learning framework improves navigation safety while maintaining strong task performance, and generalizes across different environments, robot platforms, and dynamic scenarios.

%% file: Chapters/Conclusion.tex
\section{Conclusion}

In this work, we revisit the problem of learning from demonstration for robot navigation and identify a key limitation: conventional approaches rely primarily on successful trajectories and lack awareness of safety-critical failures. This limitation often leads to unsafe behaviors when the robot encounters states that are not covered by expert demonstrations. To address this issue, we propose a failure-aware learning framework that explicitly decouples the roles of success and failure data. Instead of treating all data uniformly, the proposed method leverages failure data to shape value estimation while restricting policy learning to successful demonstrations. This structured and asymmetric design enables effective utilization of failure experiences without introducing undesirable behaviors into the learned policy.

Extensive experiments in both simulation and real-world environments demonstrate that the proposed method significantly reduces collision rates while maintaining strong navigation performance. Moreover, the learned policy generalizes well across different environments, robot platforms, and dynamic scenarios, highlighting the effectiveness of the proposed approach. This work demonstrates that failure experiences can serve as an effective source of implicit supervision for safe robot navigation when properly incorporated. The results provide empirical evidence that structured utilization of failure experiences is critical for achieving both safety and performance in learning from demonstrations.